\documentclass[10pt,twocolumn,letterpaper]{article}

\usepackage[pagenumbers]{cvpr} 

\usepackage{graphicx}
\usepackage{amsmath}
\usepackage{amssymb}
\usepackage{booktabs}
\usepackage{pbox}
\usepackage{epstopdf}
\usepackage{xspace}
\usepackage{comment}
\usepackage{lipsum}
\usepackage{bm}
\usepackage{bbm}
\usepackage{tabularx}
\usepackage{setspace}
\usepackage{sidecap}
\usepackage{xcolor}
\usepackage{wrapfig}
\usepackage{multirow}
\usepackage{array}
\usepackage{blindtext}
\usepackage{enumitem}
\usepackage{diagbox}
\usepackage{url}
\setenumerate[1]{itemsep=0pt,partopsep=0pt,parsep=\parskip,topsep=5pt}
\setitemize[1]{itemsep=0pt,partopsep=0pt,parsep=\parskip,topsep=5pt}
\setdescription{itemsep=0pt,partopsep=0pt,parsep=\parskip,topsep=5pt}
\usepackage[pagebackref,breaklinks,colorlinks]{hyperref}

\usepackage[capitalize]{cleveref}
\crefname{section}{Sec.}{Secs.}
\Crefname{section}{Section}{Sections}
\Crefname{table}{Table}{Tables}
\crefname{table}{Tab.}{Tabs.}

\newcommand{\dataset}{HOI4D\xspace}

\def\cvprPaperID{1950} 
\def\confName{CVPR}
\def\confYear{2022}

\begin{document}

\title{HOI4D: A 4D Egocentric Dataset for Category-Level Human-Object Interaction}


\author{
Yunze Liu\textsuperscript{ 1,3},
\quad Yun Liu\textsuperscript{ 1},
\quad Che Jiang\textsuperscript{1},
\quad Kangbo Lyu\textsuperscript{1},
\quad Weikang Wan\textsuperscript{2},\\
\quad Hao Shen\textsuperscript{2},
\quad Boqiang Liang\textsuperscript{2},
\quad Zhoujie Fu\textsuperscript{1},
\quad He Wang\textsuperscript{2},
\quad Li Yi\textsuperscript{ 1,3}\\
\textsuperscript{1} Tsinghua University,
\quad\textsuperscript{2} Peking University,
\quad\textsuperscript{3} Shanghai Qi Zhi Institute\\
\href{http://hoi4d.top/}{http://hoi4d.top/}
}

\maketitle

\begin{abstract}
We present \dataset, a large-scale 4D egocentric dataset with rich annotations, to catalyze the research of category-level human-object interaction. \dataset consists of 2.4M RGB-D egocentric video frames over 4000 sequences collected by 4 participants interacting with 800 different object instances from 16 categories over 610 different indoor rooms. Frame-wise annotations for panoptic segmentation, motion segmentation, 3D hand pose, category-level object pose and hand action have also been provided, together with reconstructed object meshes and scene point clouds. With \dataset, we establish three benchmarking tasks to promote category-level HOI from 4D visual signals including semantic segmentation of 4D dynamic point cloud sequences, category-level object pose tracking, and egocentric action segmentation with diverse interaction targets. In-depth analysis shows \dataset poses great challenges to existing methods and produces huge research opportunities.
\end{abstract}

\section{Introduction}
Tremendous progress~\cite{cheng2021mivos,hehefan2021p4transformer,yi2021asformer,weng2021captra} has been made for naming objects and activities in images, videos or 3D point clouds over the last decade facilitated by significant dataset and benchmark efforts. However, these perception outcomes fail to satisfy the needs of more and more critical applications such as human-assistant robots and augmented reality where the perception of interactions from 4D egocentric sensory inputs (e.g., temporal streams of colored point clouds) is required. It becomes highly desirable for a computer vision system to build up a detailed understanding of human-object interaction from an egocentric point of view. Such understanding should unify semantic understanding of 4D dynamic scenes, the 3D pose of human hands under object occlusion, the 3D pose and functionality of novel objects of interaction interest, as well as the action and intention of humans, which pose new challenges for today's computer vision systems.

\begin{figure}[t]
    \centering
    \includegraphics[width=\linewidth]{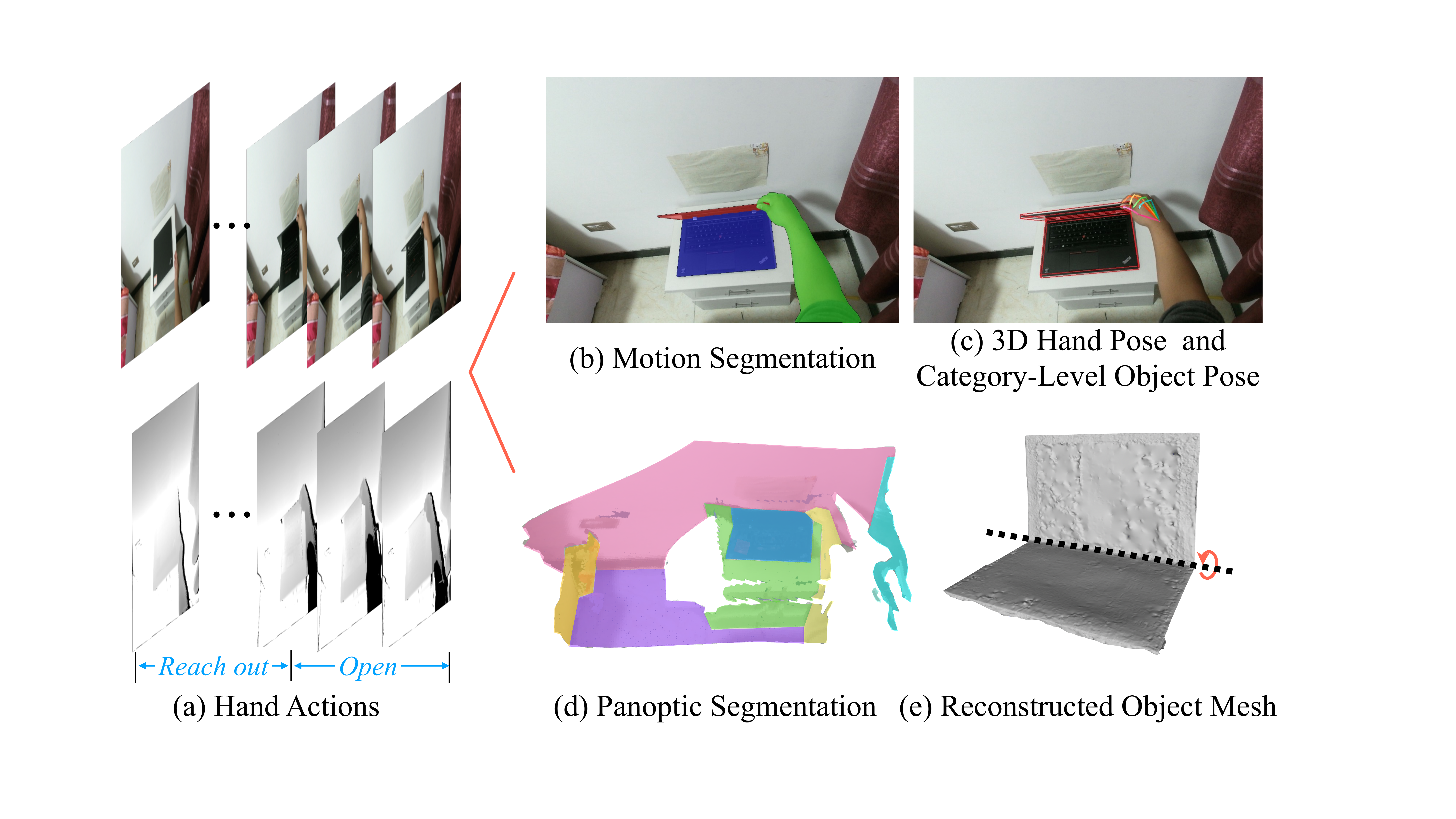}
    \caption{\textbf{Overview of HOI4D: }We construct a large-scale 4D egocentric dataset with rich annotation for category-level human-object interaction. Frame-wise annotations for action segmentation\textbf{(a)}, motion segmentation\textbf{(b)}, panoptic segmentation\textbf{(d)}, 3D hand pose and category-level object pose\textbf{(c)} are provided, together with reconstructed object meshes\textbf{(e)} and scene point cloud.}
    \label{fig:teaser}
    \vspace{-15pt}
\end{figure}

To help tackle these challenges, large-scale and annotation-rich 4D egocentric HOI datasets as well as the corresponding benchmark suites are strongly needed. Recently some efforts\cite{kwon2021h2o,garcia2018first,hampali2020honnotate} have been made to fulfill such needs. However, most of these works focus on what we called \textit{instance-level human-object interaction} where the objects being interacted with all come from a very small pool of instances whose exact CAD models and sizes are known beforehand. This impedes their application to perceiving human interaction with the vast diversity of objects in our daily life. Moreover, these works tend to ignore articulated objects while only focusing on rigid objects with which the interaction patterns are relatively simpler. These limitations partially come from the challenging and cumbersome nature of jointly capturing hands, objects and real scenes in an egocentric manner. Curating synthetic datasets~\cite{hasson2019learning} might be an alternative. Nonetheless simulating natural human motion and functional grasping for generic objects are still open research problems, making it hard for existing synthetic datasets to reach a realism sufficient for sim-to-real transfer.

To cope with the above limitations, we present, for the first time, a large-scale 4D egocentric dataset for \textit{category-level human-object interaction} as depicted in Figure~\ref{fig:teaser}. We draw inspirations from recent category-level object pose estimation and pose tracking works~\cite{wang2019normalized,li2020categorylevel,weng2021captra} and aim to propel 4D HOI perception to a new era to handle category-level object variations in cluttered scenes. We collect a richly annotated 4D egocentric dataset, \dataset, to depict humans interacting with various objects while executing different tasks in indoor environments. \dataset consists of 2.4M RGB-D egocentric video frames over 4000 sequences of 9 participants interacting with 800 object instances. These object instances are evenly split into 16 categories including both rigid and articulated objects. Also instead of using a lab setting like most previous works, the camera wearers execute tasks revealing the functionality of each category without wearing any markers in 610 different indoor scenes. \dataset is associated with reconstructed scene point clouds and object meshes for all sequences.
\dataset provides annotations for frame-wise panoptic segmentation, motion segmentation, 3D hand pose, rigid and articulated object pose, and action segmentation, delivering unprecedented levels of detail for human-object interaction at the category level.

The rich annotations in \dataset also enable benchmarking on a series of category-level HOI tasks. In this paper, we focus on three tasks in particular: semantic segmentation of 4D dynamic point cloud sequences, category-level object pose tracking for hand-object interaction and egocentric hand action recognition with diverse interaction targets. We provide an in-depth analysis of existing approaches to these tasks. Our experiments suggest that \dataset has posed great challenges to today's computer vision algorithms.
For category-level object and part pose tracking task, most of the previous datasets use synthetic data under simple scenes without hand occlusion. With the help of the proposed HOI4D dataset, researchers can now work on this more challenging task with real-world data. Due to a lack of annotated indoor datasets, semantic segmentation of 4D point clouds has been studied mainly for autonomous driving applications. HOI4D introduces more challenges such as heavy occlusion, fast ego-motion, and very different sensor noise patterns. Fine-grained action segmentation of video can help AI better comprehend interaction, but we found that existing coarse-grained methods cannot directly process fine-grained data well.

In summary, our contributions can be listed below:
\begin{itemize}
    \item We present the first dataset, \dataset, for 4D egocentric category-level human-object interaction. \dataset covers richly annotated 4D visual sequences captured while humans interact with a large number of object instances. The huge category-level object variations allow perceiving human interactions with potentially unseen objects.
    \item We present a data collection and annotation pipeline combining human annotations with automatic algorithms, effectively scaling up our dataset.
    \item We benchmark on three category-level HOI tasks covering 4D dynamics scene understanding, category-level object pose tracking and hand action segmentation. We provide a thorough analysis of existing methods and point out new challenges \dataset has posed.
\end{itemize}

\section{Related Work}
\subsection{Egocentric Human-Object Interaction Datasets}
Understanding human-object interaction has long been a pursuit for computer vision researchers and many previous efforts have been focusing on constructing third-person view datasets~\cite{chao2021dexycb, brahmbhatt2020contactpose, simon2017hand, taheri2020grab}. Recently we observe a surge of interest in perceiving human-object interaction from an egocentric view. A lot of these datasets focus on recognizing daily activities~\cite{fathi2011learning,li2015delving,bambach2015lending,damen2018scaling,pirsiavash2012detecting} and provide mostly 2D features omitting 3D annotations such as 3D hand poses and object poses, which are crucial for a comprehensive understanding of the underlying interactions.

Annotating 3D hand poses and object poses together is not an easy task though due to reciprocal occlusions. Some existing works leverage magnetic sensors or mocap markers to track 3D hand poses and object poses~\cite{garcia2018first,yuan2017bighand2,brahmbhatt2020contactpose}. However, the attached markers might hinder natural hand motion and bias the appearance of hands and objects. Other works leverage carefully calibrated multi-camera systems~\cite{chao2021dexycb, simon2017hand} or optimization algorithms~\cite{hampali2020honnotate} to ease the difficulty but are usually restricted to a third-person point of view. Most relevant to ours are a recent egocentric HOI dataset named H2O~\cite{kwon2021h2o}. They collect egocentric RGB-D videos with annotations for 3D hand poses, instance-level object poses, and action labels. However, H2O is restricted to instance-level human-object interaction covering interactions with only 8 object instances.
In addition, previous works only focus on rigid objects while we also consider articulated objects where richer interactions could happen. 

As shown in Table~\ref{tab:dataset}, we are the first to present a large-scale 4D egocentric dataset for category-level HOI covering both rigid and articulated object categories with an unprecedented level of richness in annotations.

\begin{table*}[h!]
\centering
\scriptsize
\caption{Comparison of existing HOI Datasets.}
\addtolength{\tabcolsep}{-3pt}
\label{tab:dataset}
{
\begin{tabular}{@{}>{\centering}p{0.13\textwidth}|cc>{\Centering}m{0.9cm}cccccccc>{\Centering}m{0.2cm}>{\Centering}m{1.4cm}>{\Centering}m{0.9cm}>{\Centering}m{1.7cm}}
\toprule 
dataset &4D & real& markerless &3D hand&6D obj &ego &\#frames&\#obj&\#seq&dynamic grasp& action label & seg & category-level & articulated & functional intent\\
\midrule
$\text{GTEA}_{\text{GAZE}}$+\cite{li2015delving}  &$\color{red}\times$ & \greencheck & \greencheck &$\color{red}\times$&$\color{red}\times$ &\greencheck &778K & - &37&\greencheck & \greencheck & $\color{red}\times$ & $\color{red}\times$ & \greencheck & $\color{red}\times$\\
EPIC-KITCHEN\cite{damen2018scaling}& $\color{red}\times$ & \greencheck & \greencheck & $\color{red}\times$ & $\color{red}\times$ & \greencheck & 20M & - & 700 &\greencheck & \greencheck & \greencheck & \greencheck & \greencheck & \greencheck\\
FPHA\cite{garcia2018first} & \greencheck& \greencheck& $\color{red}\times$& \greencheck& \greencheck& \greencheck & 105K & 4 & 1,175 & \greencheck & \greencheck & $\color{red}\times$ & $\color{red}\times$ & $\color{red}\times$ & \greencheck\\
ObMan\cite{hasson2019learning} &\greencheck & $\color{red}\times$& - &\greencheck&\greencheck & $\color{red}\times$ & 154K & 3K & - & - & - & \greencheck & \greencheck & $\color{red}\times$ &$\color{red}\times$\\
FreiHAND\cite{zimmermann2019freihand} &$\color{red}\times$ & \greencheck&\greencheck &\greencheck&$\color{red}\times$ &$\color{red}\times$ & 37K & 27 & - & - & $\color{red}\times$ & \greencheck & $\color{red}\times$ &$\color{red}\times$ & $\color{red}\times$\\
ContactPose\cite{brahmbhatt2020contactpose} & \greencheck & \greencheck& $\color{red}\times$& \greencheck& \greencheck&$\color{red}\times$ & 2,991K & 25 & 2,303 & $\color{red}\times$ & $\color{red}\times$ & $\color{red}\times$& $\color{red}\times$ & $\color{red}\times$ & \greencheck\\
HO-3D\cite{hampali2020honnotate} &\greencheck & \greencheck & \greencheck &\greencheck&\greencheck & $\color{red}\times$ &78K &10& 27 &\greencheck & $\color{red}\times$ & $\color{red}\times$ & $\color{red}\times$ & $\color{red}\times$ & $\color{red}\times$\\
DexYCB\cite{chao2021dexycb} & \greencheck& \greencheck& \greencheck& \greencheck& \greencheck& $\color{red}\times$& 582K & 20 & 1,000 & \greencheck & $\color{red}\times$ & $\color{red}\times$ & $\color{red}\times$ & $\color{red}\times$ & $\color{red}\times$ \\
H2O\cite{kwon2021h2o}& \greencheck & \greencheck & \greencheck & \greencheck & \greencheck & \greencheck &571K& 8 & - &\greencheck& \greencheck & $\color{red}\times$ & $\color{red}\times$ & $\color{red}\times$ & \greencheck\\
Ours &\greencheck & \greencheck & \greencheck & \greencheck & \greencheck  & \greencheck & 2.4M & 800 & 4K &\greencheck& \greencheck & \greencheck & \greencheck & \greencheck & \greencheck \\
\bottomrule
\end{tabular}

}
\end{table*}


\subsection{4D Dynamic Scene Understanding}
4D Dynamic Scene Understanding is important since it enables AI to understand the real world we live in. Existing methods are mainly based on outdoor datasets such as Synthia 4D\cite{ros2016synthia} and SemanticKITTI\cite{aygun20214d}. MinkowskiNet\cite{choy20194d} proposes to use 4D Spatio-Temporal ConvNets to extract 4D features. MeteorNet\cite{liu2019meteornet} takes point cloud sequences as inputs and aggregates information in the temporal and spatial neighborhoods. SpSequenceNet\cite{shi2020spsequencenet} manipulates the 4D point cloud data in the 3D cube style to reduce spatial information loss.
PSTNet\cite{fan2020pstnet} proposes a point spatio-temporal convolution to achieve informative representations of point cloud sequences. P4transformer\cite{fan2021point} is a novel Point 4D Transformer to avoid point tracking. 
4D-Net\cite{piergiovanni20214d} proposes a novel learning technique for fusing information in 4D from multi-modalities. 
In the indoor interactive scene, the scale of the object is smaller, the movement is more diverse, and there is even the deformation of the object, which presents new challenges to existing methods. 
\subsection{Category-Level Object Pose Estimation and Pose Tracking}
To define the poses of novel objects, NOCS\cite{wang2019normalized} proposes a Normalized Object Coordinate Space as a category-specific canonical reference frame. Every input object pixel is projected into the category-level canonical 3D space. ANCSH\cite{li2020category} extended the concept of NOCS to articulated objects and proposes Normalized Part Coordinate Space (NPCS), which is a part-level canonical reference frame. In terms of pose tracking, 6-PACK\cite{wang20206} tracks a small set of keypoints in RGB-D videos and estimates object pose by accumulating relative pose changes over time.  CAPTRA\cite{weng2021captra} build an end-to-end differentiable pipeline for accurate and fast pose tracking for both rigid and articulated objects. BundleTrack\cite{wen2021bundletrack} proposes a novel integration method and a memory-augmented pose graph optimization for low-drift accurate 6D object pose tracking. However, existing methods do not consider the pose tracking for the hand and the object jointly, which is very important in interactive scenes. Replacing the third angle of view with the ego-centric view, the problem of object occlusion becomes more serious, which also makes this task more difficult. In addition, existing datasets such as NOCS\cite{wang2019normalized} are synthetic datasets, so the domain gap between real-world data and synthetic data also poses challenges to existing algorithms. With the help of the proposed HOI4D dataset, researchers can now work on the above more challenging tasks with little overhead.

\section{Constructing HOI4D}

\subsection{Hardware Setup and Data Collection}
\begin{figure}[h]
    \centering
    \includegraphics[width=\linewidth]{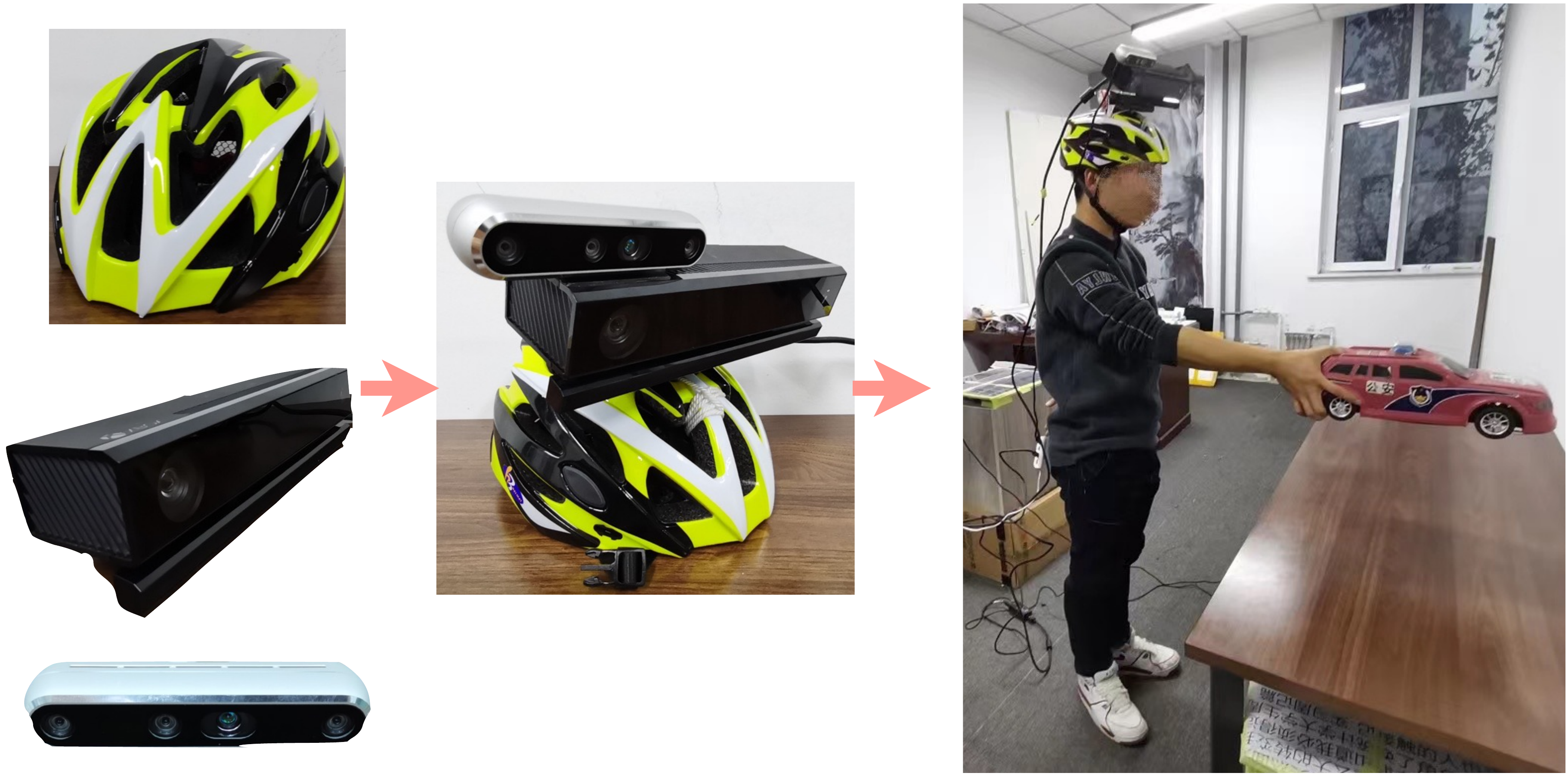}
    \caption{\textbf{Data capturing system.}we build up a simple head-mounted data capturing suite consists of a bicycle helmet, a Kinect v2 RGB-D sensor, and an Intel RealSense D455 RGB-D sensor.}
    \label{fig:capture}
    \vspace{-10pt}
\end{figure}
To construct \dataset, we build up a simple head-mounted data capturing suite consisting of a bicycle helmet, a Kinect v2 RGB-D sensor, and an Intel RealSense D455 RGB-D sensor as shown in Figure~\ref{fig:capture}. Two RGB-D sensors are pre-calibrated and synchronized before the data capturing process. Participants wear helmets to execute various tasks and interact with a diverse set of daily objects in different indoor scenes. We pre-define the pool of tasks that involves not only simple pick-and-place but also other functionality-oriented tasks such as placing a mug inside a drawer. To complete these tasks, participants need to properly plan their actions based on the specific scene configuration. For example, if the drawer is open then the participants just need to place the mug inside the drawer directly. Otherwise, they might need to open the drawer first.
Worth mentioning, we adopt two popular RGB-D cameras, Intel RealSense D455 and Kinect v2, which are good complements to each other. Kinect v2 is based upon Time of Flight (TOF) and captures long-range content. The RealSense D455 is a structured light-based camera and has more advantages within a short-range (about 1m). Together the two sensors could capture the 3D scene more comprehensively and they also provide a natural testbed for cross-sensor transfer learning.

\subsection{Data Annotation Pipeline}

\begin{figure*}[t!]
    \centering
    \includegraphics[width=0.9\linewidth]{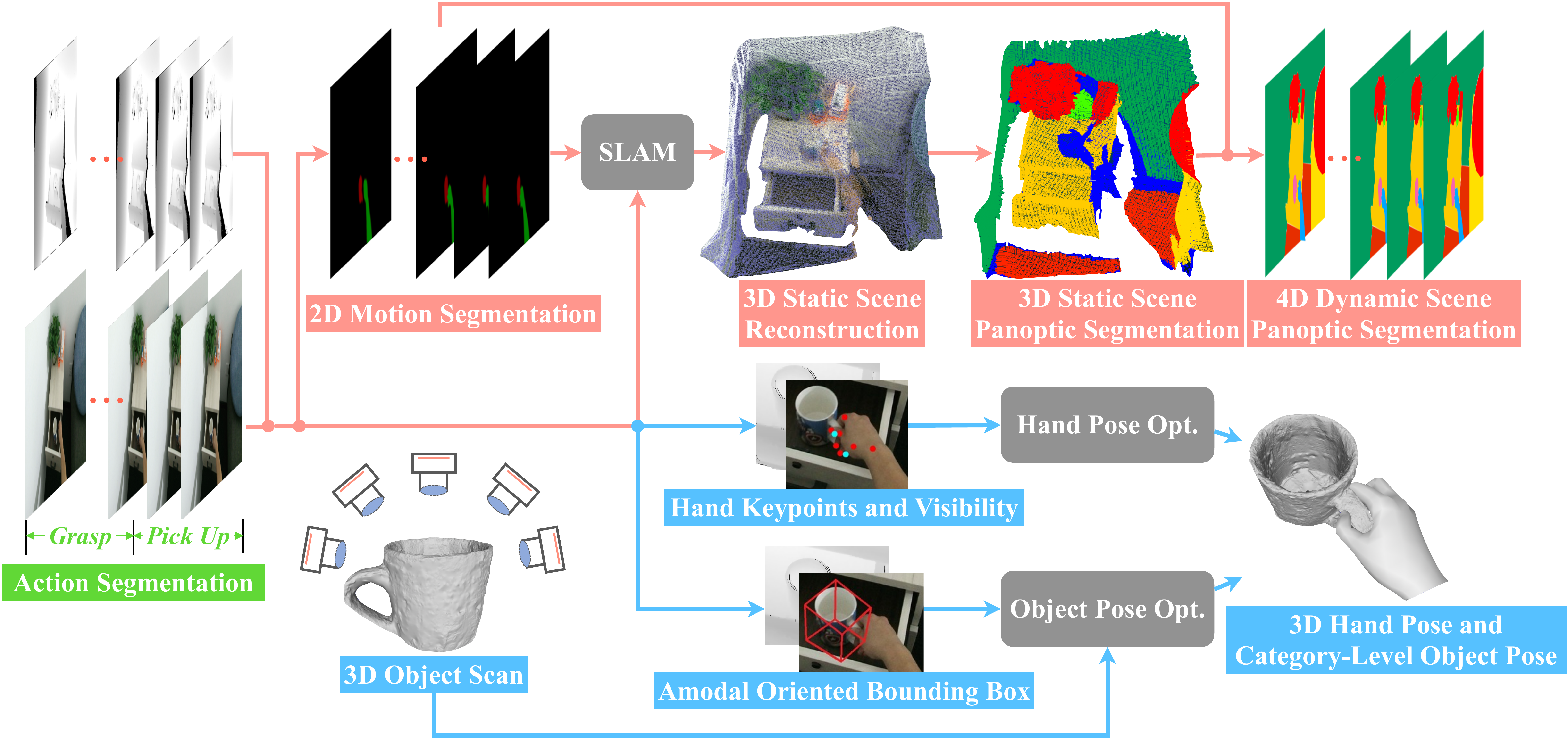}
    \caption{\textbf{Overview of annotation pipeline. }\textbf{Red branch}: Given a dynamic RGB-D sequence, we first annotate frame-wise 2D motion segmentation. Then we mask out the moving content and reconstruct a 3D static scene. We manually annotate the reconstructed scene to obtain 3D static scene panoptic segmentation. Finally, the 2D motion segmentation and the 3D static scene panoptic segmentation are merged, resulting in the 4D dynamic scene panoptic segmentation. \textbf{Blue branch}: To obtain 3D hand pose labels, we first annotate a set of hand keypoints on RGB-D frames and then leverage an optimization module to recover the underlying 3D hand. For category-level object poses, we manually fit amodal oriented bounding boxes to objects or object parts in RGB-D frames and further optimize it by leveraging the object mesh. \textbf{Green branch}: We directly annotated fine-grained action labels on the original video.}
    \label{fig:annotation}
    \vspace{-10pt}
\end{figure*}

\dataset consists of rich labels covering different aspects of category-level human-object interaction and collecting these annotations is not a trivial task. We show our data annotation pipeline in Figure~\ref{fig:annotation}. Given a dynamic RGB-D sequence, we first split moving content and static content to ease panoptic labeling by annotating framewise 2D motion segmentation. Then we mask out the moving content and reconstruct a 3D static scene via a SLAM algorithm~\cite{zhou2018open3d,choi2015redwood}. This allows us to efficiently annotate all the static content in the whole sequence. We manually annotate the reconstructed scene to obtain 3D static scene panoptic segmentation. Finally, the 2D motion segmentation and the 3D static scene panoptic segmentation are merged, resulting in the 4D dynamic scene panoptic segmentation. We explain the detailed process in Section~\ref{sec:4dsemantic}. To obtain 3D hand pose labels, we first annotate a set of hand keypoints on RGB-D frames and then leverage an optimization module to recover the underlying 3D hand as described in Section~\ref{sec:handpose}. To obtain category-level object poses, we manually fit amodal oriented bounding boxes to objects or object parts in RGB-D frames and make sure the pose definitions are consistent across a certain object category. We further optimize the object poses by leveraging the object mesh reconstructed from a multi-view scanning process. The category-level object and part pose annotation process is described in Section~\ref{sec:objpose}. Moreover, we also describe the process for action annotation in Section~\ref{sec:action_anno}.

\subsection{4D Panoptic Labeling}
\label{sec:4dsemantic}

Our 4D panoptic labeling process is mainly divided into two parts named 2D motion segmentation labeling and 3D static scene segmentation labeling.
In the process of 2D motion segmentation labeling, given an original RGB video, the annotators sample 10\% of frames evenly from a video and then manually annotate segmentation masks of the objects that have moved in the video. While manually annotating the entire video frames is labor-consuming, we instead use the off-the-shelf 2D mask propagation tool\cite{cheng2021mivos} to propagate the existed manually annotated masks to other 90\% frames in the same video. The annotators interact with the propagation tool to refine all of the segmentation masks until they are accurate.
In the process of static scene segmentation labeling, given an original RGB-D video with 2D motion segmentation masks, we remove all of the masked objects from single frame point clouds and then reconstruct the static point cloud~\cite{zhou2018open3d,choi2015redwood} of the indoor scene containing only objects whose position hasn't been changed. We label all the object instances and background stuff in the static scene and project the results back to each frame so that the segmentation masks of static objects are acquired. The final 4D panoptic segmentation labels are obtained by merging the motion segmentation and static segmentation masks.
\subsection{Hand Pose Annotation}
\label{sec:handpose}
The entire hand pose annotation process includes four stages: annotation, initialization, propagation and refinement.

Annotators uniformly annotate 20\% frames in each video. For annotation, marker-based annotation methods are unfeasible since we need realistic appearances of hands. Alternatively, we manually annotate 2D positions of a set of hand keypoints. We adopt the pre-defined 21 keypoints of hand joints that are widely used in previous works\cite{romero2017embodied,hampali2020honnotate,kwon2021h2o}. For each annotated frame, annotators provide the 2D position of 11 keypoints: the wrist, 5 fingertips and the second knuckles counted from tips. The reasonable positions of occluded keypoints are also estimated.

We represent hand pose by the MANO parametric hand model\cite{romero2017embodied}. The shape parameters $\beta \in \mathbb{R}^{10}$ are fixed based upon the real hand information from the data capturer and we optimize hand pose $\theta_h=\{\theta_m, R, t\} \in \mathbb{R}^{51}$ that consists of the pose coefficient $\theta_m \in \mathbb{R}^{45}$(3 DoF for each of the 15 hand joints) plus global rotation $R \in SO(3)$ and translation $t \in \mathbb{R}^{3}$. In the initialization stage, we estimate 3D hand poses from 2D annotations by minimizing loss functions for every annotated frame. The loss function is defined as:
\begin{equation}
\hat{\theta}_h = \mathop{\arg\min}_{\theta_h}(\lambda_j\mathcal{L}_{j}+\lambda_{2D}\mathcal{L}_{2D}+\lambda_{d}\mathcal{L}_{d}+\lambda_{pc}\mathcal{L}_{pc}+\lambda_{m}\mathcal{L}_{m})
\end{equation}
where $\mathcal{L}_{j}$, $\mathcal{L}_{2D}$,  $\mathcal{L}_{d}$, $\mathcal{L}_{pc}$ and $\mathcal{L}_{m}$ represents joint angle loss, 2D joint loss, depth loss, point cloud loss and mask loss respectively. $\lambda_j$, $\lambda_{2D}$ ,$\lambda_d$,$\lambda_{pc}$  and $\lambda_m$ are balancing parameters. Details about loss terms' definitions are given in the supplementary material.
Considering the temporal consistency of the video, each frame is initialized by the hand pose of its previous frame to accelerate convergence.
We propagate the hand poses $\theta_h$ of the annotated frames across the whole sequence by linear interpolation to obtain the coarse hand pose of each frame in the video.

In the refinement stage, we further optimize $\theta_h$ to get precise poses for all frames. The loss function is defined as:
\begin{equation}
\begin{split}
\hat{\theta}_h = \mathop{\arg\min}_{\theta_h}(\lambda_j&\mathcal{L}_{j}+\lambda_{d}\mathcal{L}_{d}+\lambda_{pc}\mathcal{L}_{pc}+\lambda_{m}\mathcal{L}_{m}\\&+\lambda_{Contact}\mathcal{L}_{Contact}+\lambda_{tc}\mathcal{L}_{tc})
\end{split}
\end{equation}
where $\mathcal{L}_{Contact}$ and $\mathcal{L}_{tc}$ represents contact loss and temporal consistency loss respectively. More details about loss terms' definition are given in the supplementary material.
To balance the efficiency and limited computational resource, we select 6-11 consecutive frames as an optimization batch containing 2-3 annotated frames.
We manually detect failure frames since the optimization process may fail due to ambiguous hand poses or a bad initialization. The hand poses of failure frames are manually rectified.
\subsection{Category-Level Pose Annotation}
\label{sec:objpose}
The process of annotating category-level poses for both rigid and articulated objects contains three stages: object measurement and annotation, model scanning and reconstruction and label propagation and pose optimization.

\noindent\textbf{Object measurement and annotation.} To balance the annotation quality and the labor intensity, we manually annotate tight amodal oriented bounding boxes for objects of interest every ten frames in each video.
Specifically, we first measure each object of interest physically and define its coordinate system to get its tight amodal bounding box. Then the annotators will manually rotate and place these boxes to fit objects in the depth point cloud of each frame.
Together with the box scales, we finally get 9D poses for rigid objects of interest. For articulated objects, we treat each part as an individual object and annotate them separately.\\
\noindent\textbf{Model scanning and reconstruction.}
Labeling object poses in 3D is a challenging task for human annotators while not all annotations are fully reliable. In addition, we only manually annotate 1 out of 10 frames so that some label propagation and optimization techniques such as~\cite{hampali2020honnotate} are needed to produce labels for all video frames. Existing pose optimization techniques are mostly designed for instance-level pose annotation which requires corresponding CAD models for objects of interest. Therefore we have scanned all the 800 objects in our dataset for pose optimization purposes.
We have covered a variety of object categories with varying sizes, materials, and topologies, making object scanning non-trivial as well.
Although a commercial 3D scanner can be used to model objects of small size, modeling objects with large size, especially with complex topology and materials remains challenging.
We choose to first manually decorate objects with various stickers that enrich the object texture and hide some of the highly specular areas. We then use off-the-shelf software packages \cite{agisoft,realitycapture} to reconstruct the object mesh from multi-view high-resolution color images. Specifically, we take images from various depression angles to entirely cover the outer surface of an object. We then adopt a series of algorithms provided by the software packages to automatically align the images, reconstruct the object mesh and calibrate model specifications. For articulated objects, we additionally provide part annotations similar to PartNet\cite{mo2019partnet}.
Worth mentioning, by providing object meshes, \dataset could facilitate research in instance-level HOI and also makes it possible to transfer the human interaction trajectories to a simulation environment for applications such as robot imitation learning~\cite{qin2021dexmv}.\\
\noindent\textbf{Label propagation.}
To propagate manually annotated object poses to intermediate frames, we convert all object poses into a world coordinate system using the camera matrix. Then we linearly interpolate the translation, rotation, and joint angles between annotated object poses.\\
\noindent\textbf{Pose optimization.}
We leverage multi-modal data, including RGB-D images, the reconstructed object mesh, as well as the 2D motion segmentation masks, to reduce the error caused by label propagation. We
utilize a differentiable renderer SoftRas\cite{liu2019soft} and auxiliary loss terms in HOnnotate\cite{hampali2020honnotate} to optimize the pose by gradient descent.  The object pose $\theta_o$, which consists of rotation $R\in SO(3)$, translation $t\in \mathbb{R}^3 $, and joint angles 
$\theta\in \mathbb{R}^{\text{\#\ of\ joints}} $ of articulated objects, should minimize the loss function defined as:
\begin{equation}
    \hat{\theta}_o = \mathop{\arg\min}_{\theta_o}(\lambda_{2D}\mathcal{L}_{2D}+\lambda_{d}\mathcal{L}_{d}+\lambda_{cd}\mathcal{L}_{cd}+\lambda_{tc}\mathcal{L}_{tc}).
\end{equation}
where $\mathcal{L}_{2D}$ and $\mathcal{L}_{d}$ are computed by SoftRas\cite{liu2019soft}, which penalize the silhouette discrepancy between the rendered object mask and the ground truth mask acquired in \ref{sec:4dsemantic}, and the depth residual between the cropped real depth map and the depth map of rendered reconstructed mesh respectively. The auxiliary loss terms are defined similar to Honnotate\cite{hampali2020honnotate}, where $\mathcal{L}_{cd}$ refers to the chamfer distance between the reconstructed object point cloud and the collected object point cloud, and $\mathcal{L}_{tc}$  maintains the temporal consistency of the pose trajectory.
For articulated objects, we additionally set the joint angle to be restricted within the physical limits of each joint.

\subsection{Action annotation}
\label{sec:action_anno}
Detecting and temporally locating action segments in long untrimmed 4D videos is a challenging task. For each frame in a video, we annotate its action category to support the study of action segmentation.
It is worth noting that we define fine-grained actions in the interactive scene, which is significantly different from the existing datasets. Detailed categories are provided in the supplementary materials.


\section{Dataset Statistics}
\begin{figure}[t]
    \centering
    \includegraphics[width=\linewidth]{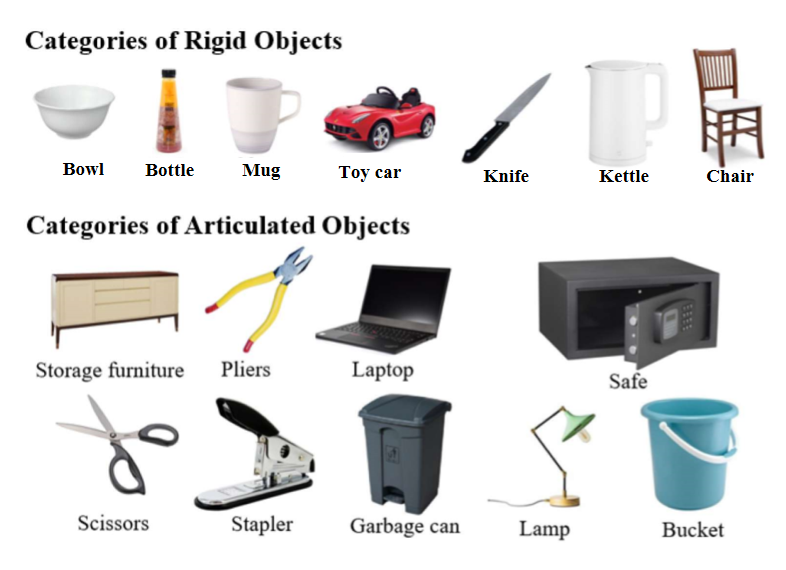}
    \caption{Diversity of object categories. }
    \label{fig:category}
    \vspace{-15pt}
\end{figure}
\noindent\textbf{Category overview.} Figure~\ref{fig:category} shows the object categories contained in our dataset. We select 16 common object categories in our daily life (7 rigid object categories, 9 articulated object categories) to construct our dataset. Each category consists of 50 unique object instances, and each object instance corresponds to a CAD model reconstructed from a set of high-resolution RGB images.
It is worth mentioning that these categories are mainly selected from ShapeNet~\cite{chang2015shapenet} and Sapien Assets~\cite{xiang2020sapien}. This makes \dataset well connected with popular synthetic 3D datasets and facilitates studying sim-to-real knowledge transfer. The reconstructed meshes and human hand trajectories can be potentially put into simulation environments to support robot learning as demonstrated in the supplementary materials.

A RealSense D455 and a KinectV2 are used to capture human-object interaction simultaneously, providing opportunities to study knowledge transfer across different depth sensors. Each video is captured at 15fps for 20 seconds. As a result, \dataset contains 2.4M frames in total.
\begin{figure}[t]
    \centering
    \includegraphics[width=\linewidth]{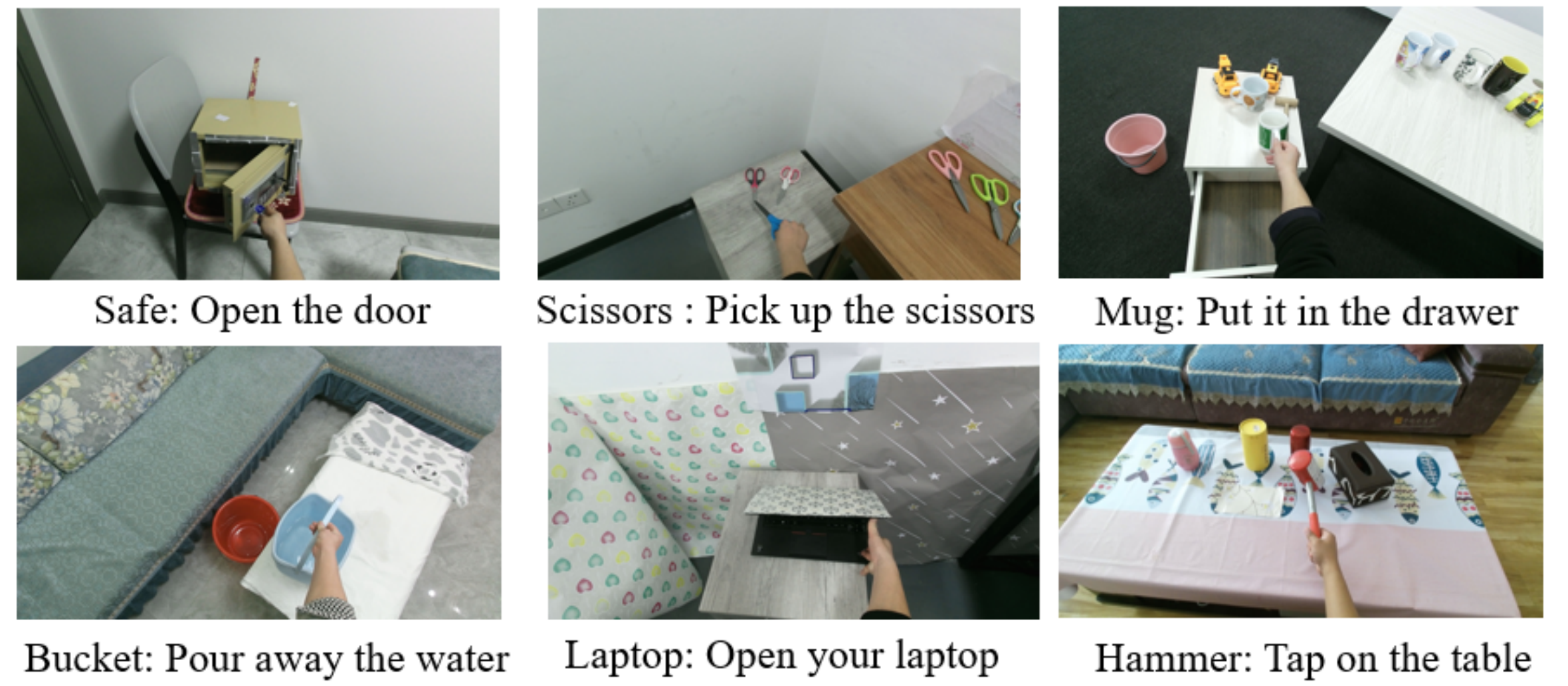}
    \caption{Examples of interaction task}
    \label{fig:task}
        \vspace{-15pt}
\end{figure}

\noindent\textbf{Diversity of interaction task.} To reflect the functionality of different object categories, we define interactive tasks based on object functions as shown in Figure~\ref{fig:task}. We have defined 54 tasks across all 16 categories. Each object includes a pick and place task and several functionality-based tasks, which can be used to support perceiving object mobility and functionality in interactive scenarios.
According to the difficulty of the tasks and the complexity of the scenes, we divide the tasks into two levels: the simple level and the complex level. For simple tasks, the captured sequences contain a subject performing a pick-and-place task over a target object with a relatively clean background regardless of the functionality of the objects. For complex tasks, we randomly pick 10-20 objects from our 800-object pool and place them in a cluttered manner. Tasks with different difficulties naturally support different research directions. Simple tasks better support research regarding pose tracking and robot learning, while complex tasks introduce interesting research problems such as 4D panoptic segmentation.

\section{Cross-Dataset Evaluations}
We conduct a cross-dataset evaluation to further assess the merit of our HOI4D dataset. We consider three tasks including 3D hand pose estimation, category-level object pose tracking and action segmentation.\\
\textbf{Settings.} For 3D hand pose estimation, we consider H2O~\cite{kwon2021h2o} with ego-centric views consistent with HOI4D, and select Mesh Graphormer~\cite{lin2021mesh} as the backbone. For category-level object pose tracking, we compare with NOCS~\cite{wang2019normalized}, the most commonly used dataset for category-level object pose tracking, evaluate on the ``bottle'' category, and choose CAPTRA~\cite{weng2021captra} as the backbone method.
For action segmentation, we choose GTEA~\cite{fathi2011learning} dataset for comparison, and evaluate with ASFormer~\cite{yi2021asformer} backbone on the 5 action classes(take, open, pour, close and put) that both datasets have.\\
\textbf{Results.} Table \ref{tab:combined} shows cross-dataset evaluation results. Take the 3D hand pose estimation for example. When we evaluate the HOI4D trained model on H2O, we observe an error increase of $2.2\times$ (from 22.3 to 48.9) due to domain gaps. However, when we evaluate the H2O trained model on HOI4D, a more severe error increase of $3.5\times$ happens, suggesting models trained on H2O generalize much worse than on HOI4D. While training on a combined set of HOI4D and H2O, we can further reduce the error from only training on H2O. But the error on HOI4D increases compared with only training on HOI4D. This suggests HOI4D complements H2O better than the opposite.
Similar conclusions can be drawn from the results of other subtasks that HOI4D is more challenging with more diverse data which yields stronger generalizability.
\begin{table}[t]
 \centering
 \small
 \begin{tabular}{l|ccc}
 \hline
 \backslashbox{test}{train} & HOI4D & H2O & HOI4D+H2O \\
 \hline
 HOI4D & \textbf{22.3} & 70.4 & 24.3 \\ 
 H2O & 48.9 & 19.9 & \textbf{15.9} \\
 \hline
 \hline
 \backslashbox{test}{train} & HOI4D & NOCS & HOI4D+NOCS \\
 \hline
 HOI4D & 55.3& 34.2& \textbf{57.1}\\
 NOCS & 50.4&70.5 & \textbf{83.7}\\
 \hline
 \hline
 \backslashbox{test}{train} & HOI4D & GTEA & HOI4D+GTEA \\
 \hline
 HOI4D & \textbf{52.3} &7.9 & 48.6\\
 GTEA & 14.1& 77.4& \textbf{90.4}\\
 \hline
 \end{tabular}
 \vspace{-2mm}
 \caption{\small Cross-dataset evaluation results. (a) Top: 3D hand pose estimation. Results are in root-relative MPJPE(mm), \textbf{lower} is better. (b) Middle: Category-level object pose tracking. Results are in 5°5cm accuracy, \textbf{higher} is better. (c) Bottom: Action segmentation. Results are in frame-wise accuracy, \textbf{higher} is better.}
 \label{tab:combined}
 \vspace{-6mm}
\end{table}

\section{Tasks and Benchmarks}
In this section, we design three specific tasks on \dataset: category-level object and part pose tracking, semantic segmentation of 4D point cloud videos, and ego-centric hand action segmentation. We follow a ratio of 7:3 to randomly split our 4D sequences into training and test sets and use the annotations from Section \ref{sec:objpose}, Section \ref{sec:4dsemantic}, and Section \ref{sec:action_anno} to support these three tasks respectively. We provide results of baseline methods and in-depth analyses of existing methods and discuss the new challenges that emerged from \dataset.

\subsection{Category-Level Object and Part Pose Tracking}
Most existing 6D object pose estimation or tracking approaches assume access to an object instance’s 3D model\cite{xiang2017posecnn,aing2021instancepose}. In the absence of such 3D models, generalization for novel instances becomes very difficult. To alleviate the dependence on CAD models, category-level object and part pose tracking is a promising direction.

In this section, we benchmark the state-of-the-art category-level object and part pose tracking algorithm BundleTrack\cite{wen2021bundletrack} on \dataset.
BundleTrack\cite{wen2021bundletrack} is a generic 6D pose tracking framework for new objects that do not rely on an instance or class-level 3D models. The evaluation protocol is the same as in prior work\cite{wen2021bundletrack,weng2021captra}. A perturbed ground-truth object pose is used for initialization.
And we also provide an ICP\cite{zhou2018open3d} baseline, which leverages the standard point-to-plane ICP algorithm implemented in Open3D\cite{zhou2018open3d}.
We select 4 rigid object categories and 1 articulated object category for experiments.
The following metrics are used:
5\textdegree5cm: percentage of estimates with orientation error \textless5\textdegree and translation error \textless5cm.
\textbf{$R_{err}$}: mean orientation error in degrees.
\textbf{$T_{err}$}: mean translation error in centimeters.
\begin{table}[h]
\centering
\vspace{-5pt}
\caption{Category-Level Object Pose Tracking on 4 rigid objects.}
\vspace{-10pt}
\label{tab:Tracking}
\newcolumntype{Y}{>{\centering\arraybackslash}X}
{
\setlength{\tabcolsep}{0.2em}
\begin{tabularx}{\columnwidth}{>{\centering} m{0.25\columnwidth}|Y|Y|Y|Y}
\toprule
  ICP& toy car & mug & bottle & bowl \\
    \hline
    5\textdegree5cm & 0.7 & 1.2 &2.1& 3.2 \\
    \hline
    \textbf{$R_{err}$}&88.3&67.9 &53.9&39.4 \\
    \hline
    \textbf{$T_{err}$}&47.6 & 20.1&28.4&23.9 \\
\bottomrule
\end{tabularx}
\vspace{-10pt}
\setlength{\tabcolsep}{0.2em}
\begin{tabularx}{\columnwidth}{>{\centering} m{0.25\columnwidth}|Y|Y|Y|Y}
\toprule
  BundleTrack& toy car & mug & bottle & bowl \\
    \hline
    5\textdegree5cm & 9.7& 12.9 &19.3& 22.6 \\
    \hline
    \textbf{$R_{err}$}&21.0 &57.5 &18.16&19.37 \\
    \hline
    \textbf{$T_{err}$}& 13.9&4.1 &7.7&5.6 \\
\bottomrule
\end{tabularx}
}
\end{table}
\vspace{-10pt}
\begin{table}[t]
\centering
\caption{Category-Level Part Pose Tracking on laptop.}
\vspace{-5pt}
\label{tab:Tracking2}
\newcolumntype{Y}{>{\centering\arraybackslash}X}
{
\setlength{\tabcolsep}{0.2em}
\begin{tabularx}{\columnwidth}{>{\centering} Y|Y|Y|>{\centering} m{0.20\columnwidth}|Y|Y}
\toprule
  \textbf{ICP} &\footnotesize{keyboard} & display & \textbf{\footnotesize {BundleTrack}} & \footnotesize{keyboard} & display \\
    \hline
    5\textdegree5cm & 0.9& 1.5 &5\textdegree5cm& 24.2 & 12.2 \\
    \hline
    \textbf{$R_{err}$}&47.3& 84.2&\textbf{$R_{err}$}&8.6& 20.5 \\
    \hline
    \textbf{$T_{err}$}&19.8 & 41.2& \textbf{$T_{err}$}&6.8& 9.47\\
\bottomrule
\end{tabularx}
\vspace{-12pt}
}
\end{table}

Table~\ref{tab:Tracking} shows the results of 4 rigid objects and Table~\ref{tab:Tracking2} shows the results of the laptop category. Taking the bottle category as an example, BundleTrack can achieve an accuracy of 86.5 (5\textdegree5cm) on the NOCS\cite{wang2019normalized} dataset where data is not captured during human-object interaction and does not suffer from heavy hand occlusions. Now it only reaches 19.3 on \dataset. This performance degradation proves category-level pose tracking is indeed very challenging in real-world interactive scenarios, where data suffers a lot from the sensor noise, complex backgrounds, hand occlusions, as well as fast motions. Most previous algorithms are firstly developed using synthetic datasets. It is interesting to see how this previously successful experience could transfer to egocentric human-object interactions. No matter focusing on sim-to-real transfer or directly learning from the real data, \dataset makes it easy to follow both paths.
It is expected that more research will focus on category-level object pose tracking in real-world interactive scenarios.

\subsection{4D Point Cloud Videos Semantic Segmentation}

Semantic segmentation of 4D point cloud videos is mostly studied in autonomous driving applications previously but seldom touched in indoor scenarios due to the lack of annotated datasets. Indoor scenes are usually more cluttered with complex layouts. In addition to this, \dataset introduces more challenges such as the heavy occlusion due to the egocentric viewpoint, fast ego-motion of 6DOF, and very different sensor noise patterns compared with LiDAR point clouds, making it hard for existing outdoor 4D semantic segmentation networks to perform well.

To demonstrate this, we benchmark two representative approaches achieving state-of-the-art performance on outdoor 4D semantic segmentation tasks: PSTNet\cite{fan2020pstnet} and P4Transformer\cite{hehefan2021p4transformer}.
We provide 38 semantic categories in total from our indoor scenes and divide all categories into two category groups(objects, background). In our 4D point cloud video semantic segmentation setting, we carefully select 376 videos within 14 semantic categories(7 for objects and 7 for background). Each frame of video is sampled as 4,096 points without color. The evaluation metrics are the mean Intersection over Unions(mIoU) over each category group and all categories.
\begin{table}[h]
\centering
\vspace{-5pt}
\caption{Semantic Segmentation of 4D Point Cloud Videos.}
\vspace{-10pt}
\label{tab:Segmentation}
\newcolumntype{Y}{>{\centering\arraybackslash}X}
{
\setlength{\tabcolsep}{0.2em}
\begin{tabularx}{\columnwidth}{>{\centering} m{0.25\columnwidth}|Y|Y|Y}
\toprule 
   Method &\footnotesize{objects}  &\footnotesize{background} &$\text{mIoU}_{\text{all}}$ \\
    \hline
    PSTNet & 31.4 & 72.6 & 52.0 \\
    \hline
    P4Transformer& 44.6 & 77.7 & 61.2 \\
\bottomrule
\end{tabularx}
\vspace{-12pt}
}
\end{table}

Current outdoor methods cannot handle indoor dynamic point cloud well, especially the object categories, as shown in Table~\ref{tab:Segmentation}. We found that the existing methods perform significantly better on the background categories than the object categories since the object categories have smaller sizes, more flexible movements, and more severe occlusion problems that make the segmentation more challenging. It may be interesting to explore methods that can simultaneously capture both object geometry and background geometry in the future.
\subsection{Fine-grained Video Action Segmentation}
Recent advances in video action segmentation have accomplished promising achievements in coarse-level segmentation on the Breakfast\cite{dataset2014_breakfast}, 50 Salad\cite{dataset2013_50salad}, and GTEA datasets\cite{dataset2011_gtea}. Fine-grained video action segmentation can help AI better understand interactive actions in interactive tasks. However, few works focus on the fine-grained video action segmentation in interactive scenarios due to the lack of a large-scale fine-grained dataset.

We consider three representative and high-performance methods: MS-TCN\cite{farha2019ms}, MS-TCN++\cite{li2020ms} and Asformer\cite{yi2021asformer}. We use the I3D feature extracted according to Section 5.1 to train the network. We use the videos' temporal resolution at 15 fps, and the dimension of the I3D feature for each frame is 2048-d. The following three metrics are reported: framewise accuracy (Acc), segmental edit distance, as well as segmental F1 scores at the overlapping thresholds of 10\%, 25\%, and 50\%. Overlapping thresholds are determined by the IoU ratio.

\vspace{-5pt}
\begin{table}[h]
\centering
\caption{Video Action Segmentation.}
\vspace{-10pt}
\label{tab:Action}
\newcolumntype{Y}{>{\centering\arraybackslash}X}
{
\setlength{\tabcolsep}{0.2em}
\begin{tabularx}{\columnwidth}{>{\centering} m{0.25\columnwidth}|Y|Y|Y|Y|Y}
\toprule
   Method& Acc & Edit & F1@10 &F1@25 &F1@50 \\
    \hline
    MS-TCN&44.2 &74.7 &55.6 &47.8&31.8\\
    \hline
    MS-TCN++ & 42.2&75.8 &54.7&46.5 &30.3\\
    \hline
    Asformer&46.8 &80.3 &58.9 &51.3&35.0\\
\bottomrule
\end{tabularx}
\vspace{-5pt}
}
\end{table}

Table~\ref{tab:Action} shows the results. Unsurprisingly, performance for all three algorithms drops by a large margin from the coarse level to the fine-grained level. Take Asformer\cite{yi2021asformer} as an example, it can only achieve an accuracy of 46.8 on \dataset but can obtain 85.6 on 50Salads\cite{dataset2013_50salad}, which shows that the existing model does not perceive the most fine-grained actions very well.
\begin{figure}[!h]
    \centering
    \vspace{-5pt}
    \includegraphics[width=\linewidth]{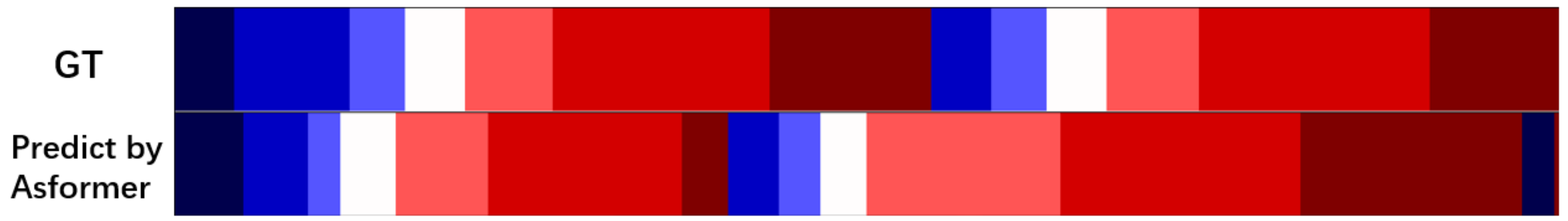}
    \caption{A qualitative analysis of failure results}
    \label{fig:video}
    \vspace{-5pt}
\end{figure}
Fig~\ref{fig:video} is an analysis of failure results: Although the prediction is completely wrong, the sequence of actions is correct. From this, we speculate that the current network learns more about the order of actions but lacks the ability to perceive the current action itself. When we use the finest-grained action labels to break the inherent sequence of actions, the performance of the existing method is greatly reduced. 

\section{Limitations and Future Work}
The main limitation of HOI4D is that human manipulation tasks with two hands are not covered since the single-hand manipulation tasks remain challenging for current research. Furthermore, the two-handed setting brings more challenges that we hope to study in the future such as cooperation of hands. As mentioned in Section~\ref{sec:objpose}, we have built a generic pipeline for creating various CAD models and the corresponding object poses of each HOI4D object category. We hope that our realistic models and rich data from HOI4D can build the bridge between simulation environments and the real world, and inspire more future research for robot learning and augmented reality applications.


{\small
\bibliographystyle{ieee_fullname}
\bibliography{egbib}
}

\clearpage
{\large
\textbf{Supplementary Material}
}

\appendix
\setcounter{section}{0}
\def\thesection{\Alph{section}}

\section{Details of HOI4D}

\subsection{CAD Model Visualization}
\begin{figure}[h]
    \centering
    \includegraphics[width=\linewidth]{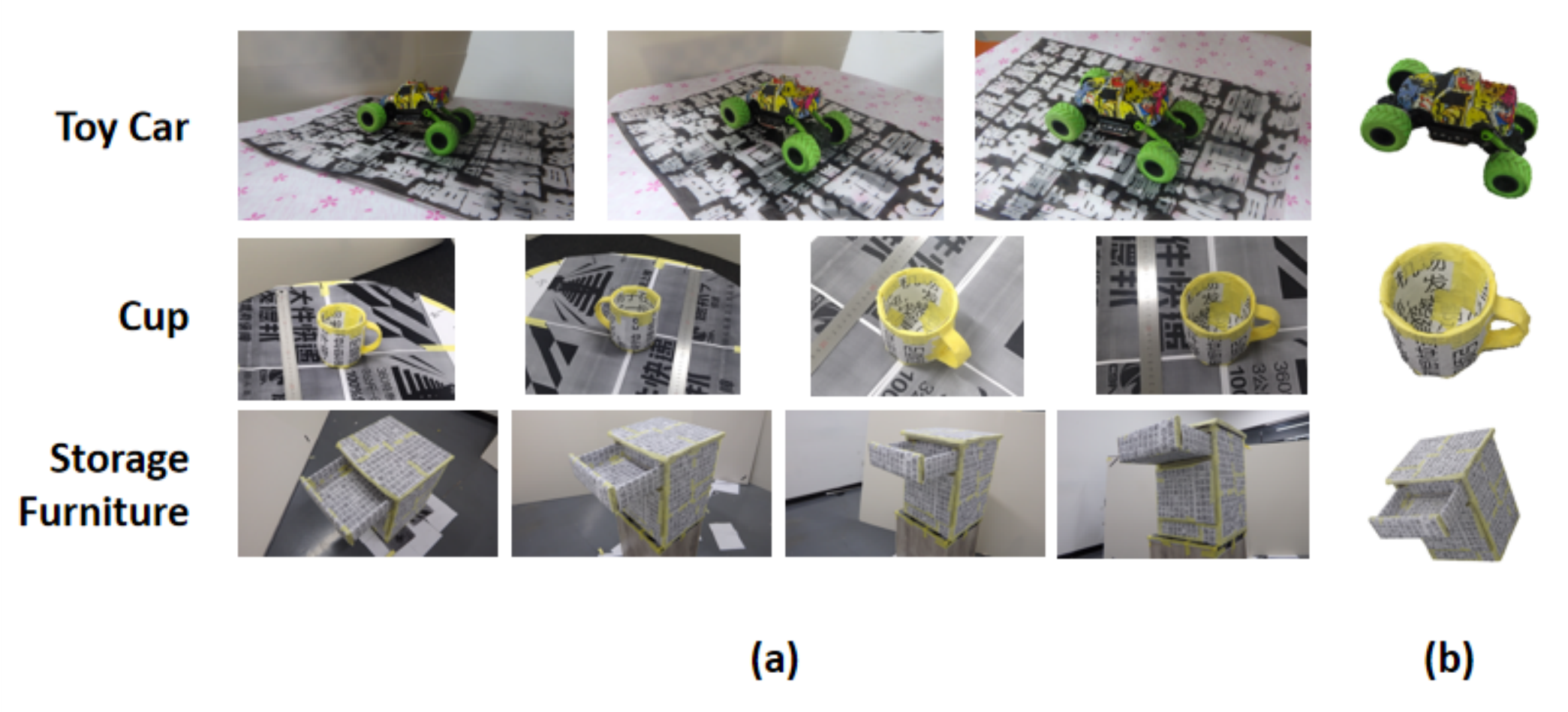}
    \caption{We show some examples of \textbf{(a)} our multi-view high-resolution color images with various depression angles, \textbf{(b)} corresponding CAD models.}
    \label{fig:supp_cad_model}
    \vspace{-15pt}
\end{figure}

\begin{figure*}[h!]
    \centering
    \includegraphics[width=\textwidth]{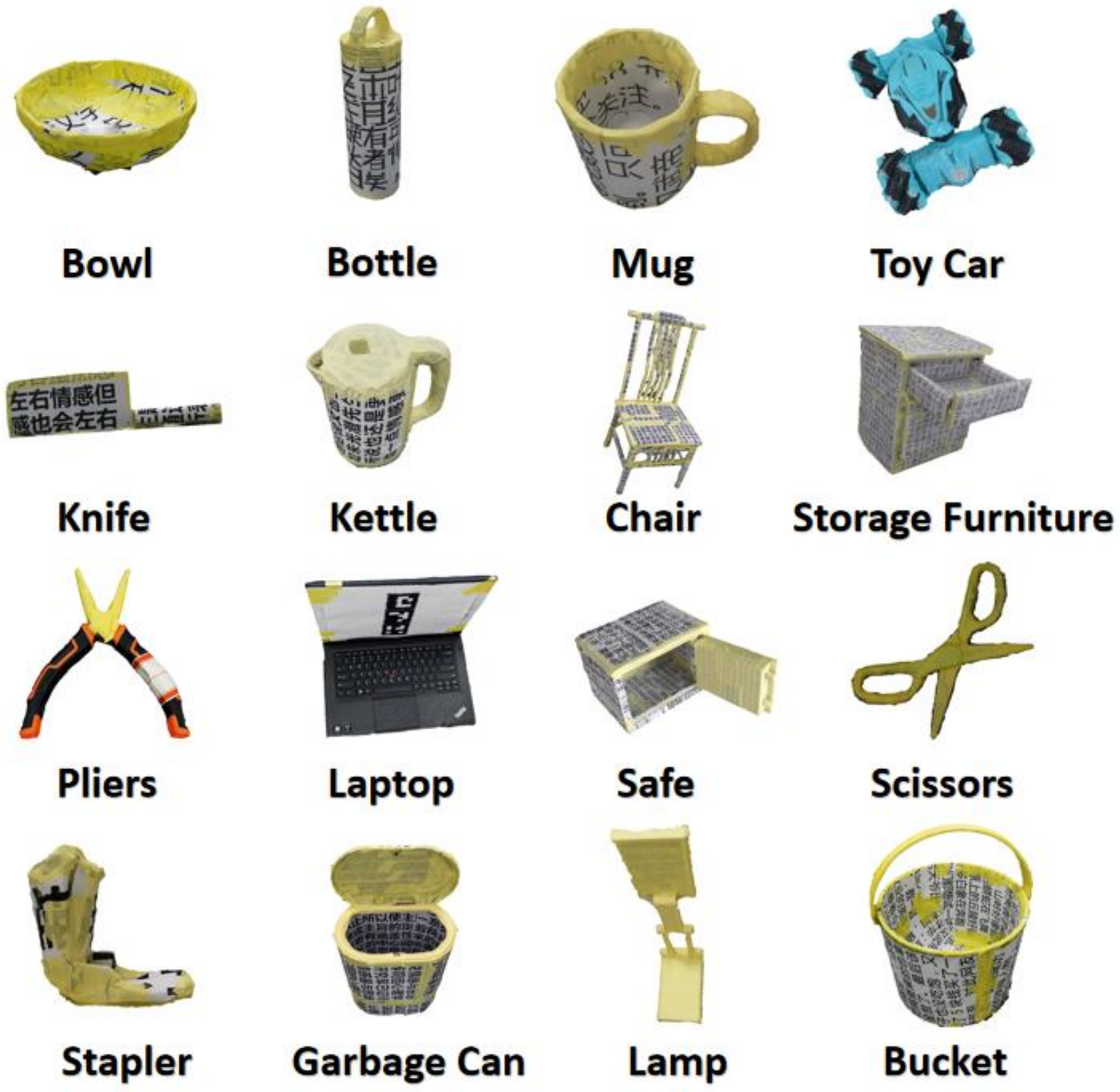}
    \caption{Some examples of CAD models among all 16 categories.}
    \label{fig:vis_cad_model}
    \vspace{-15pt}
\end{figure*}

Figure \ref{fig:supp_cad_model} shows some examples of CAD models in HOI4D. We first manually decorate objects with stickers for some categories to enrich the object texture and hide some highly specular areas. We then use off-the-shelf software packages\cite{agisoft, realitycapture} to reconstruct the CAD model from multi-view high-resolution color images. Various camera depression angles shown in Figure \ref{fig:supp_cad_model}(a) provide realistic geometry of the CAD model with the finest detail and reconstruct both inner and outer surfaces of some specific categories such as cup and storage furniture.
More visualization of CAD models in HOI4D are shown in Figure \ref{fig:vis_cad_model}.

\subsection{Dataset Statistics}

\begin{figure}[h]
    \centering
    \includegraphics[width=\linewidth]{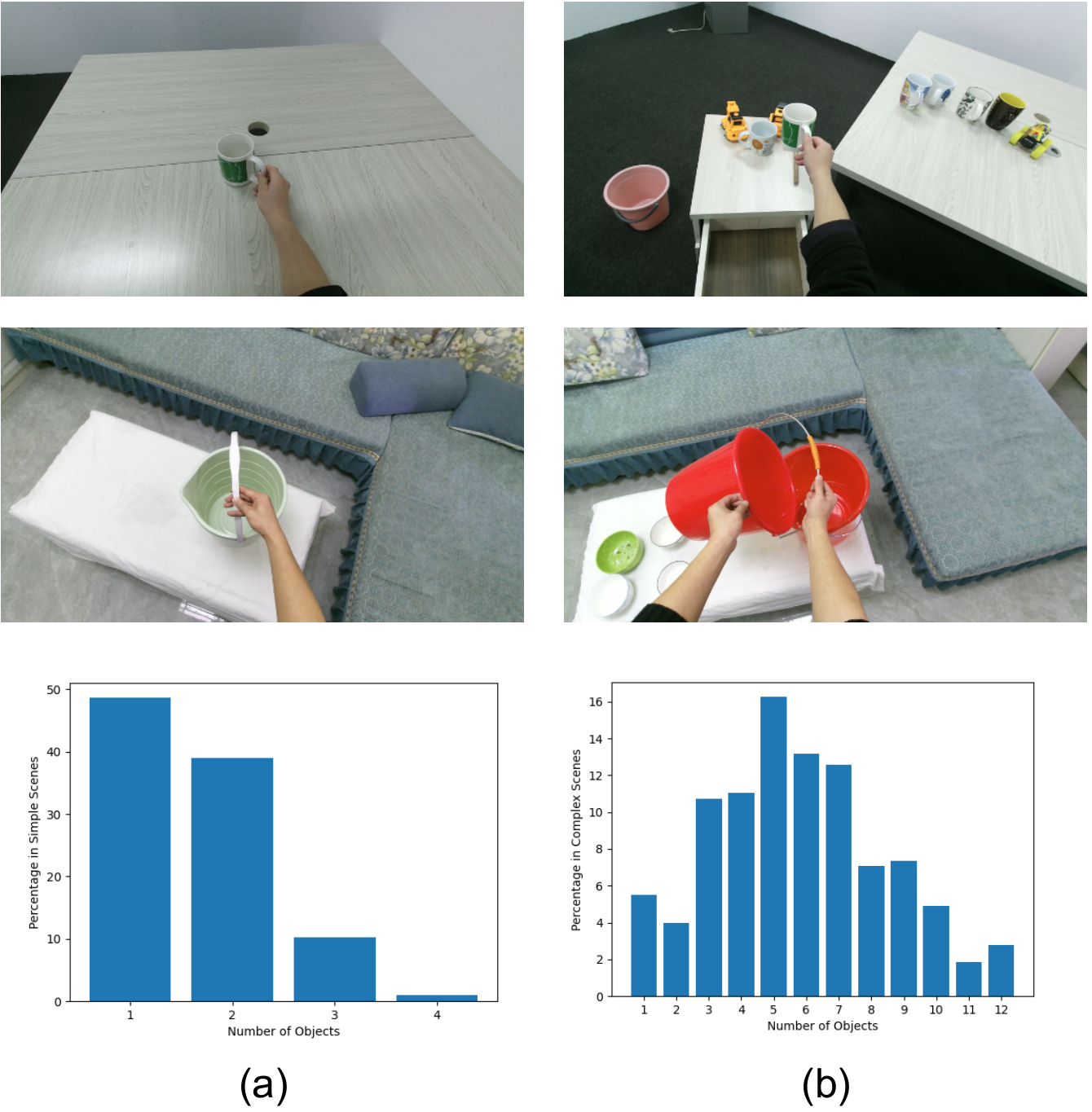}
    \caption{Sample scenes and statistics on the number of objects in a scene. \textbf{(a)} Simple scenes. \textbf{(b)} Complex Scenes.}
    \label{fig:supp_statistics}
\end{figure}

\begin{table}[h]
\centering
\newcolumntype{Y}{>{\centering\arraybackslash}X}
{
\setlength{\tabcolsep}{0.2em}
\begin{tabularx}{\columnwidth}{>{\centering} m{0.20\columnwidth}|Y}
\toprule 
   Scene &Mean (std) \\
    \hline
    Simple& 1.67 (0.79) \\
    \hline
    Complex& 6.02 (2.93) \\
\bottomrule
\end{tabularx}
\vspace{-8pt}
\caption{The average number of objects in the simple and complex scenes.}\label{tab:Statistics_1}
}
\end{table}

\begin{table}[h]
\centering
\newcolumntype{Y}{>{\centering\arraybackslash}X}
{
\setlength{\tabcolsep}{0.2em}
\begin{tabularx}{\columnwidth}{>{\centering} m{0.20\columnwidth}|Y|Y}
\toprule 
   Scene &3D static scene  &4D dynamic scene \\
    \hline
    Simple& 6.3 & 10.7 \\
    \hline
    Complex& 13.9 & 17.0 \\
\bottomrule
\end{tabularx}
\caption{The percentage of objects points in the point cloud of simple and complex scenes.}\label{tab:Statistics_2}
\vspace{-10pt}
}
\end{table}

To better support various research fields with different requirements for scene complexity, we manually divide all scenes into simple and complex scenarios. In a simple scene, the object interacted with is not obscured by surrounding objects, and the variety of camera views in the video is small to improve consistency across frames. Complex scenes have no such restrictions. Figure \ref{fig:supp_statistics} and Table \ref{tab:Statistics_1} report the number of instances in different scenarios. Moreover, Table \ref{tab:Statistics_2} reports the proportion of object points in the point cloud of scenes in diverse categories.


\subsection{Task Definitions}
Table~\ref{tab:task} shows the detailed tasks defined by each category. There are 76 tasks in all 16 categories. Pick-and-place tasks are included for each object, as well as functionality-based tasks that can be used to perceive object mobility and functionality in interactive scenarios.

\begin{table*}[h!]
  \begin{center}
    \begin{tabular}{|c|c|c|c|}
      \hline
      \textbf{Bowl} & \textbf{Bottle} &\textbf{Mug}&\textbf{Toy Car}\\
      \hline
      Pick and place & Pick and place&Pick and place& Pick and place\\
      Put it in the drawer&Pour all the water into a mug&Put it in the drawer&Push toy car\\
      Take it out of the drawer&Put it in the drawer&Take it out of the drawer&Put it in the drawer\\
      Take the ball out of the bowl&Take it out of the drawer& Fill with water by a kettle&Take it out of the drawer\\
      Put the ball in the bowl& Reposition the bottle&Pour water into another mug&\\
      Pick and place(with ball)&Pick and place(with water)&Pick and place(with water)&\\
      \hline
      \textbf{Bucket} & \textbf{Knife} &\textbf{Kettle}&\textbf{Chair}\\
      \hline
        Pick and place&Pick and place&Pick and place&Pick and place with both hands\\
      Pour water into another bucket&Put it in the drawer&Pour water into a mug&Pick and place with one hand\\
      &Take it out of the drawer&&\\
      &Cut apple&&\\
      \hline
      \textbf{Storage Furniture} & \textbf{Pliers} &\textbf{Laptop}&\textbf{Lamp} \\
            \hline
      Open and close the drawer&Pick and place&Pick and place&Pick and place\\
      Open and close the door&Put it in the drawer&Open and close the display&Turn and fold\\
      Put the drink in the door&Take it out of the drawer&&Turn on and turn off\\
      Put the drink in the drawer&Clamp something&&\\
      \hline
      \textbf{Safe} & \textbf{Garbage Can}  &\textbf{Scissors}&\textbf{Stapler}\\
            \hline
      Open and close the door&Open and close&Pick and place&Pick and place\\
      Put something in it&Throw something in it&Cut something&Bind the paper\\
      Take something out of it& &&\\
      \hline
    \end{tabular}
    \caption{Task Definitions of 16 Categories.}\label{tab:task}
  \end{center}
\end{table*}

\subsection{Hand Pose Loss Terms in Hand Pose Annotation}
\noindent\textbf{Joint angle loss.} The joint angle loss term $\mathcal{L}_{j}$ is defined following~\cite{zhou2016model} as:
\begin{equation}
\mathcal{L}_{j}=\sum_{i=1}^{45} \max(\underline{\theta_i} - \theta_m[i], 0) + \max(\theta_m[i] -
\overline{\theta_i}, 0)
\label{eq:ejoint}
\end{equation}
where $\underline{\theta_i}$ and $\overline{\theta_i}$ correspond to the lower and upper limits of the $i^{th}$ joint angle parameter $\theta_m[i]$ .\\
\noindent\textbf{2D joint loss.} The 2D joint loss of MANO joints is defined as:
\begin{equation}
\mathcal{L}_{2D} = \sum_{i=1}^{21} c[i] \Big\lVert
proj(J_{\theta_m}[i]) - J_{anno}[i] \Big\rVert^2_2
\end{equation}
where $proj(J_{\theta_m}[i])$ denotes the 2D projection of the $i^{th}$ 3D joint location $J_{\theta_m}$, $J_{anno}[i]$ is the corresponding 2D annotation, and $c[i]$ is the confidence coefficient. In practice, we set confidence coefficient $c[i]$ to 1 if joint $i$ is visible, and 0.8 otherwise. \\
\noindent\textbf{Mask loss.} The mask loss is defined following \cite{liu2019soft} as:
\begin{equation}
\mathcal{L}_{m} = 1-\frac{\lVert\hat{I}_m\otimes I_m\rVert_1}{\lVert\hat{I}_m\oplus I_m - \hat{I}_m\otimes I_m\rVert_1}
\end{equation}
where $\hat{I}_m$ and $I_m$ denote the rendered and the ground-truth 2D mask respectively. $\oplus$ and $\otimes$ are the pixel-wise product and sum operators respectively. \\
\noindent\textbf{Depth loss.} The depth loss is defined following \cite{liu2019soft} as:
\begin{equation}
\mathcal{L}_{d} = \sum_{p\in \hat{I}_m\otimes I_m} \Big\lVert
 \hat{D_p} -  D_p \Big\rVert_1
\end{equation}
where $\hat{D_p}$ and $D_p$ denote the rendered and the ground-truth depth at pixel $p$ respectively. The depth loss calculate $l_1$ loss of depth in pixels where $\hat{I}_m$ and $I_m$ intersect.\\
\noindent\textbf{Point cloud loss.} We use chamfer-distance from the rendered mano hand vertices to the ground truth point cloud which refers to the depth image cropped by 2D hand mask. This term compensates the depth loss, giving supervision to all mano hand vertices.\\
\noindent\textbf{Contact loss.} The contact loss term $\mathcal{L}_{Contact}$ is defined following \cite{hasson2019learning} as:
\begin{equation}
    \mathcal{L}_{Contact}  = \lambda_{R}\mathcal{L}_{R}+(1-\lambda_{R})\mathcal{L}_{A}
\end{equation}
where attraction loss $\mathcal{L}_{A}$ penalize hand vertices near the object's surface but are not in contact, repulsion loss $\mathcal{L}_{R}$ penalizes hand and object interpenetration. The contact weighting coefficient $\lambda_{R} \in [0,1]$ balances between $\mathcal{L}_{A}$ and $\mathcal{L}_{R}$. $\lambda_{R}=1$ when the action segmentation label indicates the hands are not interacting with the object. These terms is manually tuned for different contact modes. \\
\noindent\textbf{Temporal consistency loss.} The temporal consistency loss term $\mathcal{L}_{tc}$ is defined following~\cite{zhou2016model} as:
\begin{equation}
\mathcal{L}_{tc} = \sum_{t \in \mathcal{T}}( \lVert
\Delta_h^t \rVert^2 + \lVert \Delta_h^{t} - \Delta_h^{t-1} \rVert^2 )
\end{equation}
where $\Delta_h^t = \theta_h^t - \theta_h^{t-1}$. $\mathcal{T}$ is an optimization batch consists of 6 to 11 consecutive frames. 

\section{Qualitative analysis of Category-Level Pose Tracking}

\begin{figure}[h]
    \centering
    \includegraphics[width=\linewidth]{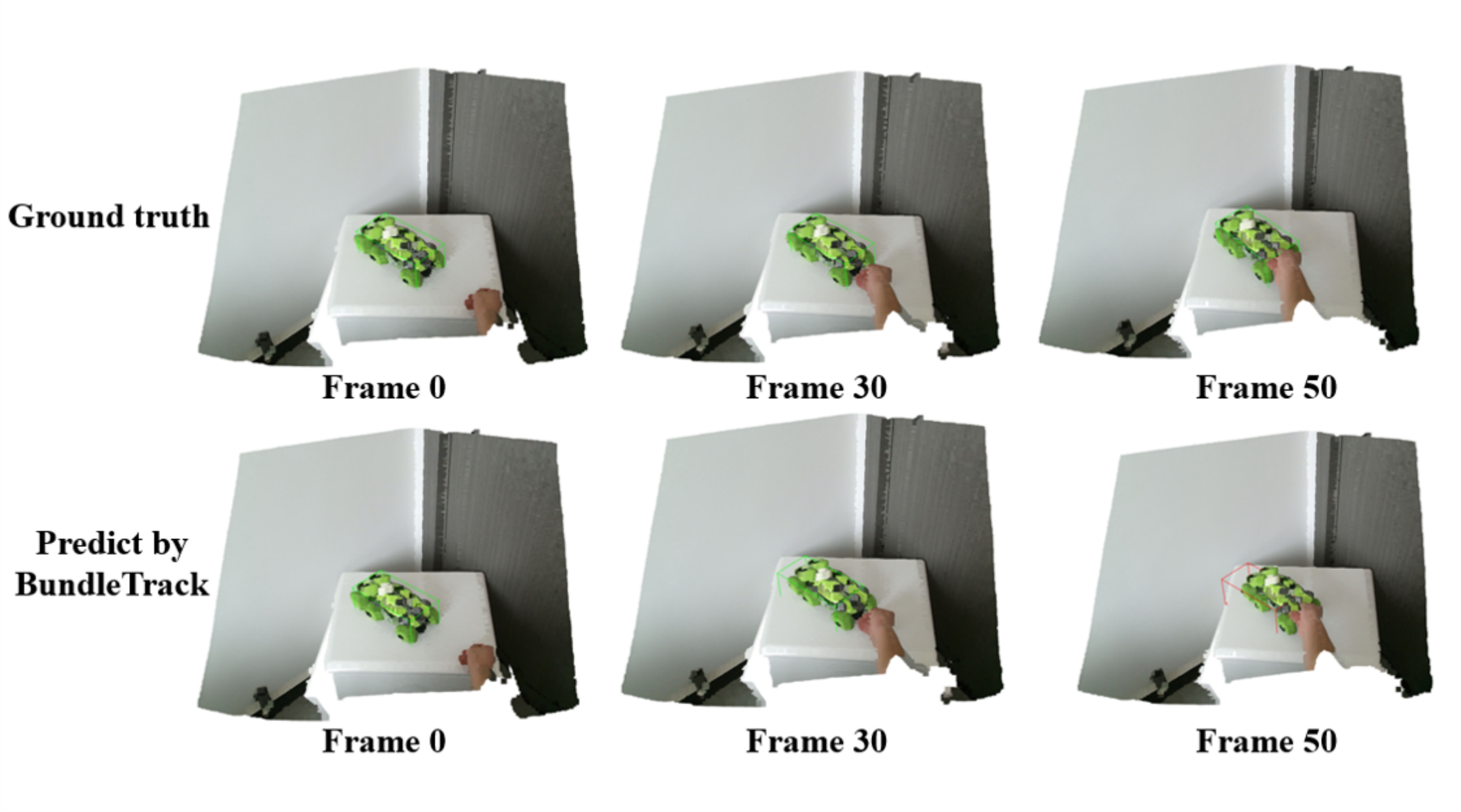}
    \caption{Qualitative evaluation on the toy car trajectory.}
    \label{fig:bundletrack1}
\end{figure}

Figure~\ref{fig:bundletrack1} below illustrates some failure cases of pose tracking experiments. We found that human objects interaction is mainly responsible for  failures for existing methods. When the camera moves, but the human is not interacting with the object, the existing methods are capable of this task; When the human interacts with the object, the occlusion of the hand and the rapid movement of the object greatly improves the difficulty of pose tracking. It is expected that more research will focus on category-level pose tracking in real-world human-object interaction scenarios.

\section{Categories of Action Segmentation}

\begin{figure}[h]
    \centering
    \includegraphics[width=0.8\linewidth]{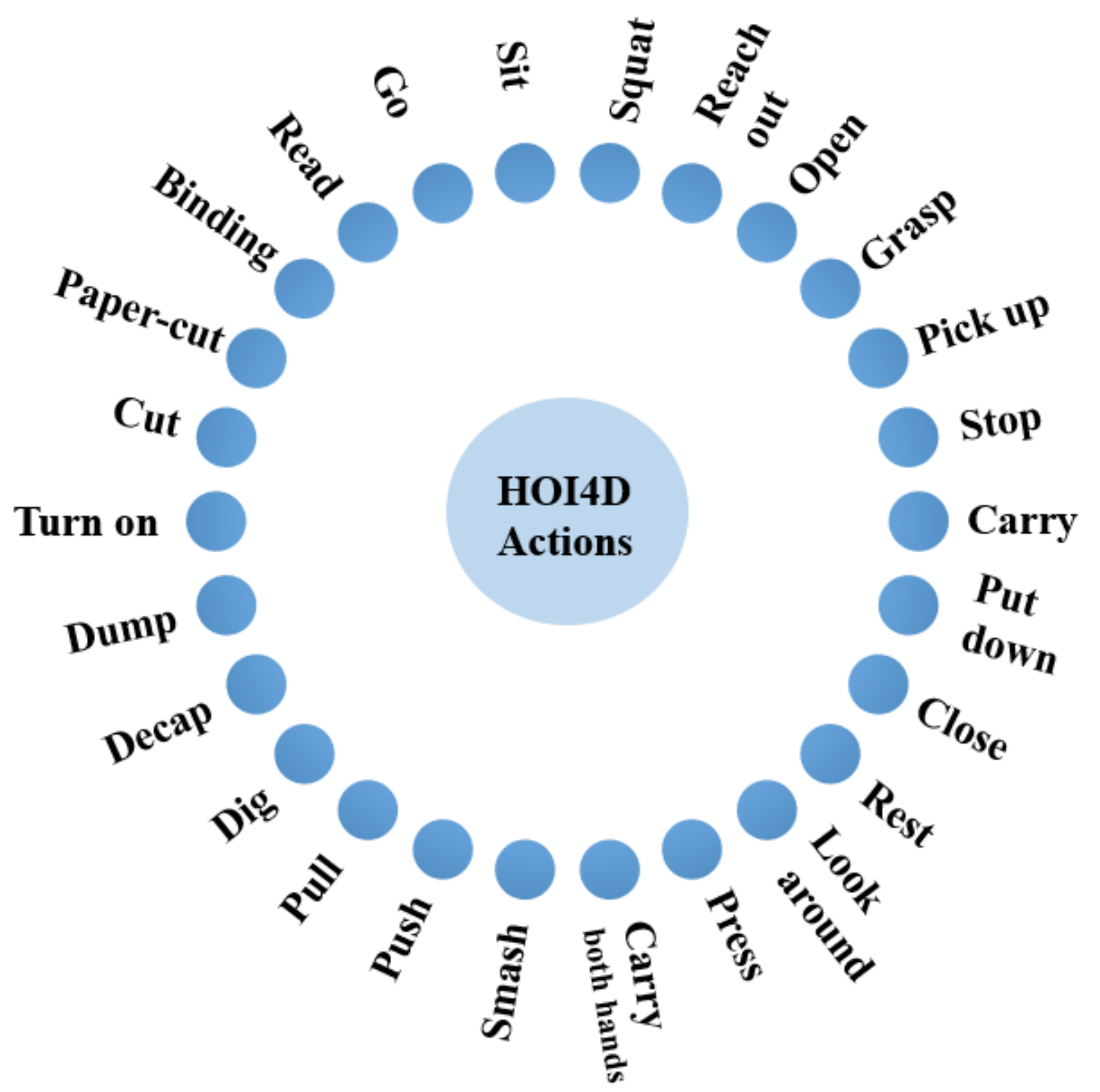}
    \caption{Categories of Action Segmentation.}
    \label{fig:actionclass}
\end{figure}
Figure~\ref{fig:actionclass} shows the action categories defined in the action segmentation task. Different from the action definition in the existing action segmentation dataset, our action category is a finer-grained label in the scene of human-object interaction.

\section{Imitation Learning for Robot Dexterous Manipulation}
HOI4D not only can serve for vision benchmarks regarding category-level human-object interaction but also provides rich knowledge for robot learning. The human-object interaction trajectories combined with the pose states of objects and hands can be naturally treated as demonstrations for robot imitation learning. This allows robots to accomplish complex interactions with various objects, which are still very challenging in the robotics community. In this section, we consider a new robotics task named category-level dexterous manipulation, where a dexterous hand needs to manipulate novel object instances from a known object category in a predefined way. Specifically, we train a robotic agent to execute an object manipulation task in a simulation environment following the design of ManiSkill\cite{mu2021maniskill}, but with the actuator being a dexterous hand. We validate that existing state-of-the-art RL algorithms could hardly achieve satisfactory results on this challenging task, while significant progress can be made through imitating the rich demonstrations in HOI4D. This shows the value HOI4D brings to the robot learning community. Below we present the experiment in detail.
\subsection{Environment Setup}
Inspired by ManiSkill\cite{mu2021maniskill}, we build an environment based on SAPIEN\cite{xiang2020sapien} simulator.\\
\textbf{Environment design:} the environment use the SAPIEN simulator with timestep set to 0.05. The environment supports various interactions between the models presented in HOI4D and PartNet and the SAPIEN model of robot Adroit Hand. In this section, we choose toy car models for imitation learning research. \\
\textbf{Task definition:} we define a \emph{Pick Up} task that requires the Adroit Hand to pick up a toy car on the table and leave it a certain height off the table. The task is successful if the toy car is close enough to the target location and is kept static for a period of time afterwards. The time limit for each episode is 200, and an episode will be evaluated as unsuccessful if it goes beyond the time limit.\\
\textbf{Observation:} the observation of the task is composed of three components: (\romannumeral1) joint angles and joint velocities of the Adroit Hand; (\romannumeral2) global position and velocities of the Adroit Hand root; (\romannumeral3) point cloud of the scene(4 dimensions: 3 for xyz position, 1 for segmentation that discriminates Adroit Hand and car). The Observation is suitable for studying category-level generalization. The point cloud is captured from eight cameras mounted on the table to provide a panoramic and object-centric view. To avoid losing too much visual information when the Adroit Hand is in contact with the object, four of these cameras are close to the object and $90^{\circ}$ apart from each other, and the other four cameras are farther away from the object. In addition, we downsample the point cloud to 1200 points to increase training speed.\\
\textbf{Action:} the action space of the task is the motor command of the 30 joints of the Adroit Hand. We use PID controllers to control the joints of the Adroit Hand. We use velocity controllers for all the joints. During training, the range of the velocities is normalized to (-1, 1). The action space corresponds to the normalized target velocities of all controllers. \\
\textbf{Reward:} the reward function of the environment is set upon three stages: (\romannumeral1) the first-stage reward gives punishment to the distance between Adroit Hand's palm and the center of the toy car. It also contains a positive proportion of the dot product of the vector from the Adroit Hand to the object and the vector with palm orientation; (\romannumeral2) the second-stage reward discourages the relative velocity between the Adroit Hand and the object. It is also negatively related to the total distance between finger tips and the object mesh; (\romannumeral3) the final-stage reward is defined based on the height of the Adroit Hand and the object. The reward is positively correlated with these heights. \\
\subsection{Demonstration Collection}
Our demonstrations for imitation learning consists of observations that represent states of the Adroit Hand and the object being interacted with and actions that represent the motion of the Adroit Hand, which are called state-action demonstrations. We transform real hand-object interaction videos in HOI4D to the state-action demonstrations in SAPIEN\cite{xiang2020sapien}, and the main challenges of this process are three-fold. First, since the manipulation tasks in the simulation environment only focus on the Adroit Hand and the object interacted with, the complicated scenarios in real videos may interfere with the imitation learning process and become noise. Second, the human hand and robot Adroit Hand are intrinsically different. Third, compared with the third-person view, egocentric videos in HOI4D suffer from more severe occlusion of both human hand and object, which brings more inaccurate human hand and object pose annotations for demonstration collection.

Inspired by DexMV\cite{qin2021dexmv}, we divide the demonstration collection process into three steps named hand joint retargeting, state-only demonstration collection and state-action demonstration collection. We transform the human hand pose represented as 51 DoF MANO model\cite{romero2017embodied} to 30 DoF Adroit Hand pose in the hand joint retargeting step. We then combine poses of Adroit Hand and object to generate state-only demonstrations, and finally compute the Adroit Hand action vectors to obtain the state-action demonstrations and use them as inputs of imitation learning.

\noindent\textbf{Hand joint retargeting.} Our hand joint retargeting method is inspired by DexMV\cite{qin2021dexmv}. Given the original 3D positions of human hand keypoints represented by MANO model\cite{romero2017embodied}, our goal is to find the optimum robot Adroit Hand pose that minimizes the distances between corresponding keypoints from human hand and Adroit Hand. We use the task space vectors\cite{handa2020dexpilot,qin2021dexmv} and accordingly define the objective function\cite{qin2021dexmv} of the optimization problem. The 15 task space vectors used in our method are designed based on DexPilot\cite{handa2020dexpilot}: vectors between five fingertips and vectors from wrist to fingertips for both human hand and Adroit Hand.

\begin{figure}[t]
    \centering
    \includegraphics[width=1\linewidth]{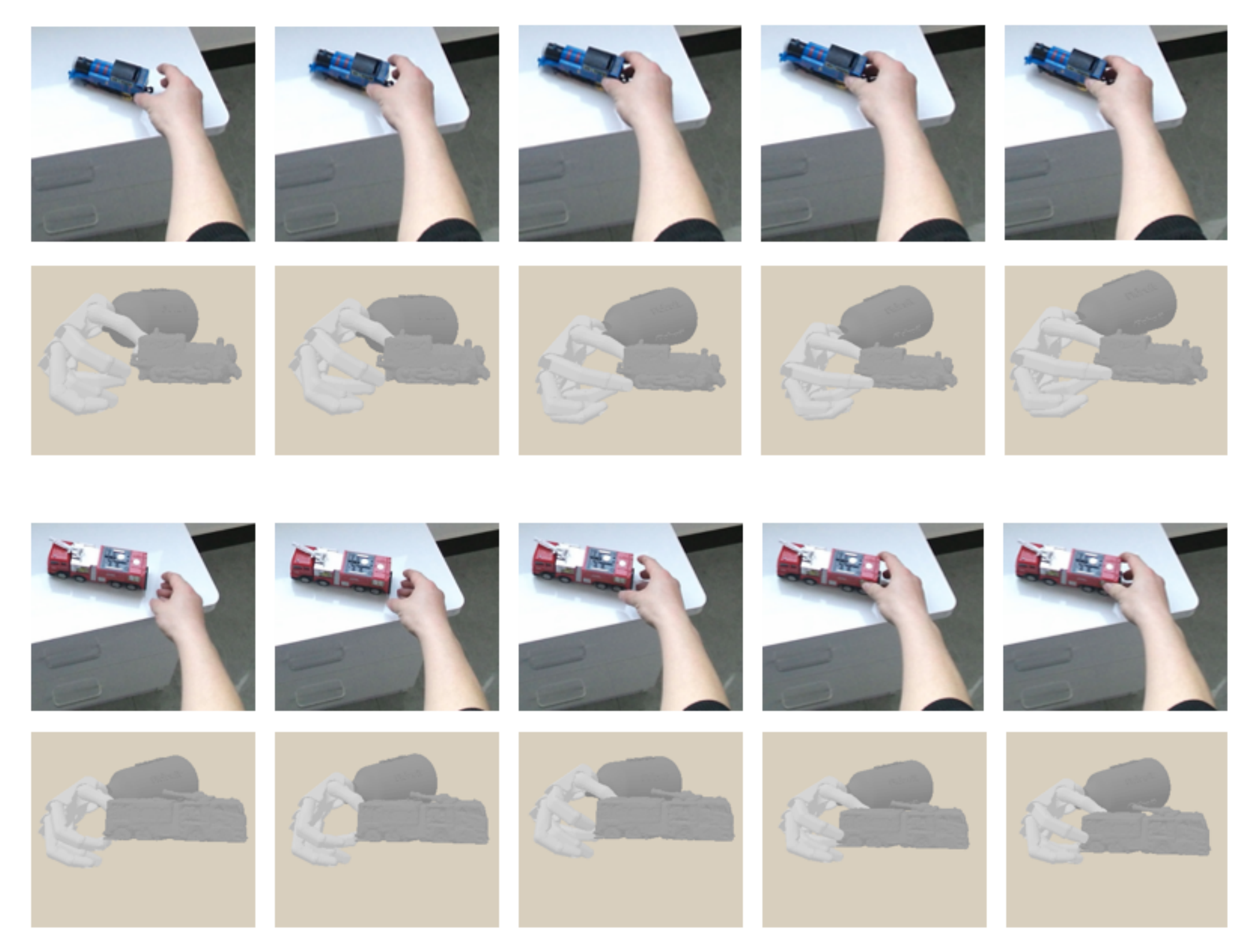}
    \caption{Examples of state-only demonstrations.}
    \label{fig:supp_demonstrations}
\end{figure}

\noindent\textbf{State-only demonstration collection.} While human hand-object interaction videos are all provided from simple scenes in HOI4D, a variety of backgrounds that are unrelated to the interaction process, such as tables and sofas with diverse geometries and positions, may increase the variance among videos and impede the learning of the hand-object interaction process. Thus we obtain the human hand and object poses using methods mentioned in the main paper, and then other information in the videos are all discarded. We use CAD models consistent with actual objects to build the bridge between real-world and simulation environments and place the CAD model and the Adroit Hand in corresponding positions and orientations according to the poses from videos. Figure \ref{fig:supp_demonstrations} shows some examples of our state-only demonstrations.

\noindent\textbf{State-action demonstration collection.} In our manipulation task settings, the action vector is directly obtained from the linear or angular velocity of each joint of the Adroit Hand. For each joint of the Adroit Hand. We fit the sequence of its angles with a cubic spline interpolation model in the time dimension to improve the smoothness of the motion trajectory, and the action can compute from the derivative of the fitting function. Combining the state-only demonstrations with the action vectors, we obtain the state-action demonstrations for imitation learning.

\subsection{Imitation Learning Approaches}

We perform imitation learning algorithms by augmenting the Reinforcement Learning algorithms using our demonstrations generated by the above methods. We compare such approaches with the RL algorithms. For RL algorithms, we adopt the commonly used Soft Actor-Critic(SAC)\cite{haarnoja2018sacapps}. For imitation learning algorithms, we adopt the Generative Adversarial Imitation Learning(GAIL)\cite{NIPS2016_cc7e2b87}, which is the SOTA IL method in robotic manipulation. For a fair comparison, we use SAC algorithm in the RL part of GAIL with the same hyper-parameters. (Other RL and IL algorithms will be studied in our future work).

We use point cloud-based vision architecture as our feature extractor since the input of our network contains point cloud. The point cloud features include position(3 dimensions: xyz)and segmentation masks(1 dimension for Adroit Hand)and we concatenate the robot state(joint angles and joint velocity of the Adroit Hand, global position and velocities of Adroit Hand root)to the features of each point.

For the RL algorithm, we parameterized the policy network and the value network with the \emph{PointNet+Transformer} architecture that is the baseline backbone in ManiSkill\cite{mu2021maniskill}.

For the IL algorithm, we use the same architecture in the policy network and value network as the RL algorithm. For the discriminator network in GAIL, we also use the \emph{PointNet+Transformer} architecture. While the expert trajectories are crucial for GAIL, we carefully select 12 demonstrations with high quality from different toy cars as our expert trajectories.
\subsection{Results and Analysis}
We evaluate the RL and IL methods on the \emph{Pick Up} task. In the experiments, the success rate is evaluated with three random seeds. The results is presented in Table \ref{tab:reselt_1}. 

\begin{figure}[t]
    \centering
    \includegraphics[width=\linewidth]{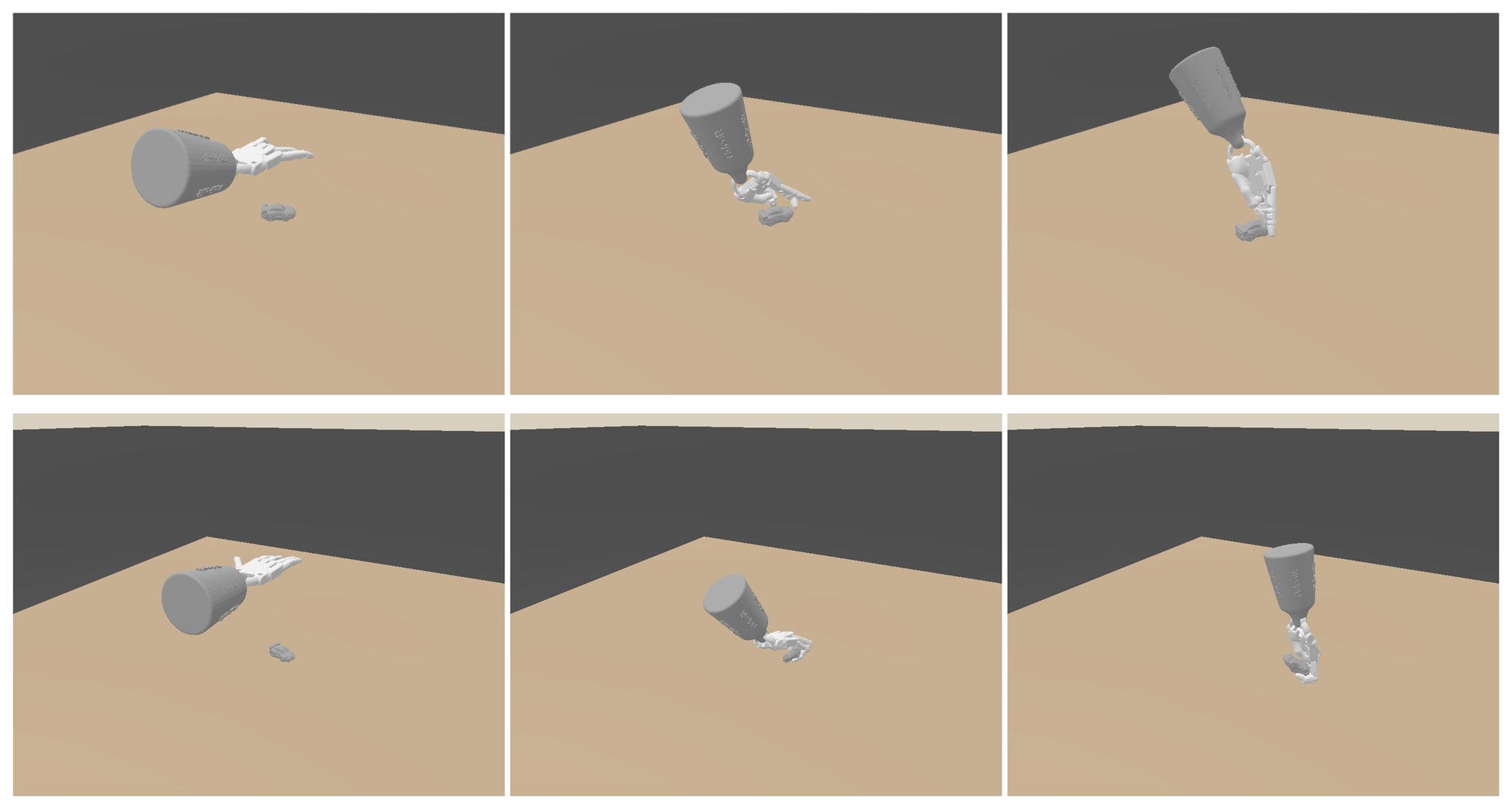}
    \caption{Comparison of RL agent and IL agent on the \emph{Pick Up} task. \textbf{(Top Row)} RL agent. \textbf{(Bottom Row)} IL agent.}
    \label{fig:RL-IL}
\end{figure}

Figure \ref{fig:RL-IL} shows a comparison of RL agent and IL agent on the \emph{Pick Up} task.

Table \ref{tab:reselt_1} shows that the IL algorithm can outperform the RL algorithm. While category-level dexterous manipulation is still a challenging task in robotics, the success rate of RL and IL algorithms are both low. 
Since our input observation for RL and IL is the point cloud of the scene, it is harder for RL and IL agents to learn to pick up the toy car than the previous works that always use the ground truth states of the environment as input observation. We use only 12 demonstrations in IL that are generated from different objects in the toy car category. These demonstrations can greatly improve the success rate of the task, which shows that our demonstrations in HOI4D can greatly help RL learn better policy.

\begin{table}[h]
\centering
\vspace{-8pt}
\newcolumntype{Y}{>{\centering\arraybackslash}X}
{
\setlength{\tabcolsep}{0.2em}
\begin{tabularx}{\columnwidth}{>{\centering} m{0.20\columnwidth}|Y}
\toprule 
   Method &Mean success rate (std)  \\
    \hline
    RL& 3.5 (3.2) \\
    \hline
    GAIL& 17.4 (7.9) \\
\bottomrule
\end{tabularx}
\vspace{-8pt}
\caption{The average success rate of different methods.}\label{tab:reselt_1}
\vspace{-10pt}
}
\end{table}

\end{document}